\documentclass[11pt]{article}

\usepackage{booktabs} 
\usepackage{natbib}
\usepackage{amsmath}
\usepackage{amsfonts}
\usepackage[ruled,vlined]{algorithm2e}
\usepackage{epstopdf}
\usepackage{graphicx}
\usepackage{xcolor}

\newcommand{\gjh}[1]{\textcolor{black}{#1}}
\newcommand{\strike}[1]{}
\newcommand{\zycone}[1]{\textcolor{black}{#1}}

\begin{document}

\title{Approximation Trees: Statistical Stability in Model Distillation}

\author{Yichen Zhou, Zhengze Zhou, Giles Hooker \\
        Department of Statistical Science\\
       Cornell University\\
       Ithaca, NY 14853, USA}
\date{}
       


\maketitle

\begin{abstract}
This paper examines the stability of learned explanations for black-box predictions via model distillation with decision trees. One approach to intelligibility in machine learning is to use an understandable ``student'' model to mimic the output of an accurate ``teacher''.  Here, we consider the use of regression trees as a student model, in which nodes of the tree can be used as ``explanations'' for particular predictions, and the whole structure of the tree can be used as a global representation of the resulting function.

However, individual trees are sensitive to the particular data sets used to train them, and an interpretation of a student model may be suspect if small changes in the training data have a large effect on it.  In this context, access to outcomes from a teacher helps to stabilize the greedy splitting strategy by generating a much larger corpus of training examples than was originally available. We develop tests to ensure that enough examples are generated at each split so that the same splitting rule would be chosen with high probability were the tree to be re-trained.  Further, we develop a stopping rule to indicate how deep the tree should be built based on recent results on the variability of Random Forests when these are used as the teacher.  We provide concrete examples of these procedures on the CAD-MDD and COMPAS data sets.
\end{abstract}

\section{Introduction}

This paper examines the use of regression trees for model distillation. While Machine Learning has traditionally focused on predictive performance, there has been considerable recent interest in ``X-raying the black box'': finding methods to make the ways in which neural networks, Random Forests and other predictive models arrive at their predictions understandable to humans. This problem can be approached by creating summaries of these models such as variable importance scores \citep{breiman2001random}, partial dependence  or ICE plots \citep{friedman2001greedy,Goldstein2013}, saliency maps \citep{simonyan2013deep} and other local explanations \citep{ribeiro2016should}. It can also be approached by developing intelligible ``student'' models which mimic the predictions of the original ''teacher'' black box: a strategy encompassed by the term {\em model distillation}.  Within model distillation,  common student models are generalized additive models (GAMS: see \citet{lou2012,tan2017detecting}, \citet{hooker2007generalized} provides a link between these and PDPs) and decision trees  \citet{breiman1984classification,quinlan1987generating}, which  are our focus.  Decision trees have an intelligible graphical representation and can automatically fit complex high-dimensional functions, both of which make them appealing as student models.  They have been used to reduce the computational cost of predictions \citep{craven1995extracting},  reduce effort in data collection (see \citet{gibbons2013computerized} especially for shortening medical questionnaires) and the final few nodes on a tree can provide an explanation for its predictions \citep{ribeiro2016should, Hu2018}. They have been used to represent complex models in \citet{johansson2010oracle,johansson2011one,he2012imitation, augasta2012reverse} and as a means of examining disparate impact in machine learning models in \citet{chouldechova2017fair}.

The features listed above make decision trees attractive as a machine learning technique. However, the greedy algorithm used to build trees results in high variability and poor performance when used directly on training data. This is because small perturbations of the data used to build the tree can result in dramatically different models as when, for example, a different covariate is chosen in a high-level split with consequences that cascade through the rest of the tree structure.  In the context of model distillation, this instability is an important concern: an explanation or interpretation of a learning outcome that is sensitive to small changes in the data may be viewed as unreliable. A related question is to determine the depth of the tree: while explanations of the global behavior of a predictive model focus on the leading nodes, those for individual predictions typically discuss the bottom of the tree. This is also where the tree structure is most variable and determining at what level of detail a teacher model needs to be explained is also relevant.

We address both these questions here. In order to obtain a stabilized structure for an approximation tree we take advantage of our ability to generate an arbitrarily large data set from which to build it.  Specifically,  we follow \citet{gibbons2013computerized} in generating pseudo data from a kernel density estimate based on the observed feature variables and using the value of the teacher model at these points as a response. In this paper we additionally ensure that, were this pseudo data to be re-generated, the same tree structure would be chosen with high probability. To carry this out, at each node we assess the stability of the selected split via a hypothesis testing framework; when splitting, we generate a large enough corpus of pseudo data to ensure that separation between the Gini index split criterion at the chosen split and that of other candidates is large enough to be consistently selected. This framework is repeated at each split to obtain a stabilized tree, generating new pseudo data as needed.

As our experiments show, this can result in the need to generate very large sets of pseudo-data when competing splits produce very similar improvement and achieving a stabilized tree can be computationally demanding. We think that this is an important observation: that many existing uses of decision trees in model distillation may produce unstable model interpretations or explanations and our understanding of these models may rest more on the particular  data used to generate the approximation tree than on the underlying structure of the teacher. There are some subtle distinctions to be made here: if a distillation tree replaces the learned model when making predictions, we might reasonably choose to present it as an explanation for how a prediction is made, even if the structure of the tree was originally determined partially by chance. However, if we also hope to interpret reasoning behind the prediction, or expect the tree to explain something about the teacher, we would require explanations to be reproducible.

Our second question concerns a stopping rule for the depth of an approximation tree: we would like to capture the structure of the teacher in so far as it relates to the underlying signal in the data, but we also want to avoid mimicking artifacts of the particular data set used to obtain it. That is, we would like to stop building a tree once nodes start capturing the variance of the teacher.  Few machine learning methods come with statistical estimates of sample variability; however for the specific case of Random Forests \citep{breiman2001random}, \citet{mentch2016quantifying} and \citet{wager2017estimation} were able to demonstrate that the predictions of Random Forests are asymptotically normally distributed with a variance that can be estimated at no additional overhead.  Here we use this to develop a test of whether the values of a Random Forests teacher model are statistically different from a constant within the current leaf of the student. This in turn allows us to annotate the student model with an assessment of whether each of the leaf nodes could yield further insight, or predictive accuracy, if split again.   Here our experiments suggest that the signal from Random Forests is generally more complex than is captured by an individual approximation tree grown to common depths, suggesting that explanations based on our approximation tree procedure are at least conservative.

In the remainder of this paper, Section \ref{sec:stabilizing}  develops a means of ensuring the stability of a single split based on ideas from hypothesis testing. This is then employed to conduct a stabilized decision tree in Section \ref{sec:stabletrees} where we also present the results of some simulation experiments. Section \ref{sec:stoppingrules} develops our stopping criterion. We then employ the complete procedure on two data sets: the CAD-MDD data set studied in \citet{gibbons2013computerized} in which the goal is to shorten a medical questionnaire screening for severe depression, and the COMPAS data set \citet{larson2016we} used to assess recidivism risk, also studied in \citet{kleinberg2016inherent,chouldechova2017fair}.  Studies on further data sets are presented in the Supplemental Material.

\section{Stabilizing Splitting Rules} \label{sec:stabilizing}

We begin by examining the statistical properties of deciding on a single split within a tree. Decision trees proceed by recursively splitting the covariate space in two based on the values of one covariate. That is, we define a split function
\[
G(X) = 1_{(X_j > c)}
\]
which divides the covariate space in two with the division indicated by the label $G(X) \in \{0,1\}$. Decision tree algorithms select the split function $G(X)$ by examining all possible splitting rules among the possible choices of covariates, $j$, and split points $c$ and choosing the values which minimize a data-fitting criterion. In this paper we focus on multi-class classification with a tree built using the change in Gini index measured over $k$ classes:
\begin{align}
g  = \sum_{i\not=j}P(Y=i)P(Y=j)  = 1-\sum_{i=1}^k P(Y=i)^2
\label{eqn:gini}
\end{align}
between the root and the two leaf nodes. In practice, this criterion is evaluated on data by replacing the probabilities in \eqref{eqn:gini} with their empirical counterparts on data.

In the context of model distillation, we consider a black box classifier $\mathcal{F}$ along with a pseudo-sample $\{(X_i, Y_i)\}_{i=1}^{n}$ of arbitrary size $n$. Here $X_i=(X_i^1,\dots,X_i^m) \in \mathbb{R}^m$, and $Y_i=(Y_i^1,\dots,Y_i^k) \in \mathbb{R}^k$ are the $\mathcal{F}$-predicted class probabilities over responses and we choose the split that minimizes a Gini index of the $Y_i$.  In order to ensure that our choice of split is stable, we need to choose $n$ large enough to control the probability that two different pseudo samples, $\{(X_i, Y_i)\}_{i=1}^{n}$ and $\{(X_i^*, Y_i^*)\}_{i=1}^{n}$ would result in different splits. Here, we make pairwise comparisons between the current best split, and the list of candidate alternatives. For each alternative, the p-value for a test that the difference in Gini gains is greater than zero gives us an estimate of the probability that a different data set would choose the alternative over the current best split. Summing these probabilities gives a bound on the likelihood of splitting the current node a different way and we then select $n$ to control this probability.

The remainder of this section provides the technical details required to carry this calculation out. We first derive a central limit theorem for the difference in Gini indices between splits; we then show how this can be used to assess the probability that a different pseudo-sample would result in a different choice of split and increment $n$ until we control this probability at a desired level. Finally, we control for multiple comparisons through a Bonferroni procedure.

\subsection{Asymptotic Distribution of Gini Indices}

A theoretical discussion of the evaluation of splits can be found in \citet{banerjee2007confidence}. In our specific case, we compare the Gini indices of candidate splits: To do so, we examine their asymptotic behavior and obtain a central limit theorem (CLT) so tests can be developed based on a normal distribution; (\ref{eqn:gini}) implies an averaging over all samples when calculating the Gini index, suggesting the existence of the CLT.

To examine two prospective splits $G_1$ and $G_2$ with the same samples, recall their Gini gains
\begin{align*}
g_1 &= 1-\pi_{1,l} \left(\sum_{j=1}^k \theta_{1,l,j}^2\right) - \pi_{1,r} \left(\sum_{j=1}^k \theta_{1,r,j}^2\right),\\
g_2 &= 1-\pi_{2,l}\left(\sum_{j=1}^k \theta_{2,l,j}^2\right) - \pi_{2,r}\left(\sum_{j=1}^k\theta_{2,r,j}^2\right),
\end{align*}
where $\pi$ represents the covariate distribution of $\tilde{X}$ and $\theta$ the conditional probability of $\hat{\tilde{Y}}$ given $\tilde{X}$. Subscripts are arranged in the order of the split, the left (denoted as $l$) or right (denoted as $r$) child, and the class label. For instance,
\begin{align*}
\pi_{1,l} &= P(G_1(X)=0), \\
\pi_{1,r} &= P(G_1(X)=1), \\
\theta_{1,l,j} &= P(Y=j|G_1(X)=0),\\
\theta_{1,r,j} &= P(Y=j|G_1(X)=1),
\end{align*}
and respectively for $G_2$. The empirical versions, Gini indices, are
\begin{align*}
\hat{g}_{1,n} &= 1-\frac{n_{1,l}}{n} \sum_{j=1}^k \left(\hat{\theta}_{1,l,j}\right)^2 - \frac{n_{1,r}}{n} \sum_{j=1}^k \left(\hat{\theta}_{1,r,j}\right)^2,\\
\hat{g}_{2,n} &= 1-\frac{n_{2,l}}{n} \sum_{j=1}^k \left(\hat{\theta}_{2,l,j}\right)^2 - \frac{n_{2,r}}{n} \sum_{j=1}^k \left(\hat{\theta}_{2,r,j}\right)^2.
\end{align*}
Moving to the left and right children of both splits, we denote the numbers of samples and the probabilities over class labels in each child by, for $p \in \{1,2\}, j \in \{1,\dots,k\},$
\begin{align*}
n_{p,l} &= \sum_{i=1}^n 1_{\{G_p(X_i)=0\}}, \\
\hat{\theta}_{p,l,j} &= \frac{1}{n_{p,l}} \sum_{i=1}^n Y_i^j\cdot 1_{\{G_p(X_i)=0\}}, \\
n_{p,r} &= \sum_{i=1}^n 1_{\{G_p(X_i)=1\}}, \\
\hat{\theta}_{p,r,j} &= \frac{1}{n_{p,r}} \sum_{i=1}^n Y_i^j\cdot 1_{\{G_p(X_i)=1\}}.
\end{align*}
For simplicity we write, for $p \in \{1,2\},q\in\{l,r\},$
$$
N_{p,q} = \begin{bmatrix}
n_{p,q}\hat{\theta}_{p,q,1}\\
\vdots \\
n_{p, q}\hat{\theta}_{p,q,k}
\end{bmatrix}, \quad
\Theta_{p,q} = \begin{bmatrix}
\pi_{p,q} \theta_{p,q,1}\\
\vdots \\
\pi_{p,q}\theta_{p,q,k}
\end{bmatrix}.
$$
Employing a multivariate CLT we obtain
\begin{align*}
\sqrt{n} \left(\frac{1}{n}\begin{bmatrix}
N_{1,l} \\
N_{1,r} \\
N_{2,l} \\
N_{2,r}
\end{bmatrix} -
\begin{bmatrix}
\Theta_{1,l} \\
\Theta_{1,r} \\
\Theta_{2,l} \\
\Theta_{2,r}
\end{bmatrix} \right) \longrightarrow
N(0, \Sigma).
\end{align*}
To relate this limiting distribution to the difference of Gini indices we shall employ the $\delta$-method. Consider the analytic function $f: \mathbb{R}^{4k} \to \mathbb{R}$ s.t.
\begin{align*}
f(x_1, \dots, x_{4k}) = &-\frac{1}{\pi_{1,l}}\sum_{i=1}^k x_i^2
-\frac{1}{\pi_{1,r}} \sum_{i=k+1}^{2k} x_i^2 \\
&+\frac{1}{\pi_{2,l}} \sum_{i=2k+1}^{3k} x_i^2
+\frac{1}{\pi_{2,r}} \sum_{i=3k+1}^{4k} x_i^2.
\end{align*}
The $\delta$-method imples that
\begin{align}
\sqrt{n} \left(f\left(\frac{1}{n}\begin{bmatrix}
N_{1,l} \\
N_{1,r} \\
N_{2,l} \\
N_{2,r}
\end{bmatrix}\right) - f\left(
\begin{bmatrix}
\Theta_{1,l} \\
\Theta_{1,r} \\
\Theta_{2,l} \\
\Theta_{2,r}
\end{bmatrix}\right)\right)
\longrightarrow N(0, \Theta^T\Sigma\Theta).
\label{eqn:main}
\end{align}
Here we write
\begin{gather*}
\Theta  = f'\left(\begin{bmatrix}
\Theta_{1,l} \\
\Theta_{1,r} \\
\Theta_{2,l} \\
\Theta_{2,r}
\end{bmatrix} \right)
= 2 \begin{bmatrix}
-\Theta_{1,l} \\
-\Theta_{1,r} \\
\Theta_{2,l} \\
\Theta_{2,r}
\end{bmatrix} \in \mathbb{R}^{4k},\\
\Sigma  = cov \begin{bmatrix}
N_{1,l} \\
N_{1,r} \\
N_{2,l} \\
N_{2,r}
\end{bmatrix} = cov \begin{bmatrix}
Y\cdot 1_{\{G_1(X)=0\}}\\
Y\cdot 1_{\{G_1(X)=1\}}\\
Y\cdot 1_{\{G_2(X)=0\}}\\
Y\cdot 1_{\{G_2(X)=1\}}
\end{bmatrix} \in \mathbb{R}^{4k \times 4k}.
\end{gather*}
We should point out that \gjh{while} (\ref{eqn:main}) provides us with the CLT we \gjh{\strike{required to qualify} need to assess} the difference between two Gini indices. After expanding (\ref{eqn:main}),
$$
\sqrt{n} \left ((\hat{g}_{1,n}-\hat{g}_{2,n}) - (g_1 - g_2)\right) \longrightarrow N(0, \Theta^T \Sigma \Theta).
$$
or asymptotically,
$$
(\hat{g}_{1,n} - \hat{g}_{2,n}) - (g_1 - g_2) \sim N \left(0, \frac{\Theta^T \Sigma \Theta}{n}\right).
$$
Hence, by replacing $\Theta, \Sigma$ by the empirical versions from the pseudo samples, we write
\begin{gather}
\hat{g}_{1,n}-\hat{g}_{2,n} \sim N\left(g_1 - g_2, \frac{\hat{\Theta}^T \hat{\Sigma}\hat{\Theta}}{n}\right).
\label{eqn:bs2}
\end{gather}
\zycone{
Denote $\Delta_n=g_{1,n}-g_{2,n}$ the Gini difference we can rewrite (\ref{eqn:bs2}) as
\begin{gather}
\hat{\Delta}_n \sim N\left(\Delta_n, \frac{\hat{\Theta}^T \hat{\Sigma}\hat{\Theta}}{n}\right).
\label{eqn:bs3}
\end{gather}
}

\subsection{The Probability of Choosing the Same Split Again}

\zycone{
The above formula (\ref{eqn:bs2}) gives rise to the following test when comparing two splits with different batches of pseudo samples. Suppose we have two  prospective splits $G_1$ and $G_2$. After drawing pseudo samples $\{(X_i, Y_i)\}_{i=1}^{n}$ and observing, without loss of generality, that $\hat{\Delta}_n = \hat{g}_{1,n} - \hat{g}_{2,n} < 0$. We intend to claim that $G_1$ is better than $G_2$. In order to ensure this split is chosen reliably, we can run a single-sided test to check whether we would obtain the same decision when accessing $\hat{\Delta}^*_n = \hat{g}^*_{1,n} - \hat{g}^*_{2,n} < 0$ with another independently-generated set of pseudo samples $\{(X^*_i, Y^*_i)\}_{i=1}^{n}$. Assuming that $\{(X_i, Y_i)\}_{i=1}^{n}$ and $\{(X^*_i, Y^*_i)\}_{i=1}^{n}$ are independent samples, (\ref{eqn:bs3}) implies
$$
\hat{\Delta}_n^* - \hat{\Delta}_n \sim N\left(0, \frac{2\hat{\Theta}^T \hat{\Sigma}\hat{\Theta}}{n}\right),
$$
which gives,
$$
\hat{\Delta}_n^* \bigg| \left(\hat{\Delta}_n = \hat{g}_{1,n} - \hat{g}_{2,n}\right) \sim N\left( \hat{g}_{1,n} - \hat{g}_{2,n}, \frac{2\hat{\Theta}^T \hat{\Sigma}\hat{\Theta}}{n}\right).
$$
}\zycone{This distribution leads to a prediction interval based on which we would get the prediction of the Gini difference using a different pseudo sample. In order to control $P(\hat{\Delta}_n^*<0)$ at a confidence level $1-\alpha$, we need
\begin{eqnarray}
\label{eqn:bs}
\hat{g}_{1,n} - \hat{g}_{2,n} < Z_{\alpha}\cdot \sqrt{\frac{2\hat{\Theta}^T \hat{\Sigma}\hat{\Theta}}{n}},
\end{eqnarray}
}where $Z_{\alpha}$ is the $1-\alpha$-quantile of a standard normal. With a sufficiently large $n$ it is always possible to determine the better split between $G_1$ and $G_2$ should they have any difference. In addition, by combining this test with a pairwise comparisons procedure, we are capable of finding the best split among multiple prospective splits.

\subsection{Sequential Testing}

The power of  this split test increases with $n$. Since we need to choose $n$ to reveal any detectable difference between two splits, when no prior knowledge is given regarding the magnitude of the difference, we need an adaptive approach to increasing $n$ accordingly.

\gjh{\strike{Provided} For a fixed} confidence level $\alpha$, suppose we have tested at sample size $n$ and get p-value $p_n > \alpha$. Referring to (\ref{eqn:bs}), we have
$$
\sqrt{n}\cdot \frac{\hat{g}_{1,n}-\hat{g}_{2,n}}{\sqrt{2\hat{\Theta}^T \hat{\Sigma} \hat{\Theta}}} = Z_{p_n}.
$$
Notice that $\dfrac{\hat{g}_{1,n}-\hat{g}_{2,n}}{\sqrt{2\hat{\Theta}^T \hat{\Sigma} \hat{\Theta}}}$ is the estimator of $\dfrac{g_1-g_2}{\sqrt{2\Theta^T \Sigma \Theta^T}}$ which is an intrinsic constant with respect to the pairwise comparison. Hence in order to reach a p-value less than $\alpha$ we may increase sample size to $n'$ such that
\begin{align*}
\sqrt{n'}\cdot \frac{\hat{g}_{1,n}-\hat{g}_{2,n}}{\sqrt{2\hat{\Theta}^T \hat{\Sigma} \hat{\Theta}}} = Z_{\alpha},
\end{align*}
which yields that
\begin{gather}
\sqrt{\frac{n}{n'}} = \frac{Z_{p_n}}{Z_{\alpha}}.
\label{eqn:ad1}
\end{gather}

Due to pseudo sample randomness, a few successive increments are required before we land in the confidence level.

\subsection{Multiple Testing}
So far we have obtained a method to compare a pair of splits. But when splitting a certain node we usually need to choose the best split among multiple $G_1,\dots G_m$. In order to adapt our pairwise  split test to this situation, we consider modifying the problem slightly into deciding whether the split with the minimal Gini index is intrinsically superior than any other splits. This problem can be resolved by conducting pairwise comparisons of the split with minimal Gini index against the rest.

If we still want to test at a certain significance $\alpha$  whether the split with the lowest estimated Gini index, i.e, $\hat{g}_{n,(1)}$, is the optimal, we can still work within the scheme of the pairwise comparison with an additional procedure controlling the familywise error rate (FWER). Here we make an analogue of the Bonferroni correction \citep{dunnett1955multiple}.

$\bullet$ Test the hypotheses $H_{i,0}: g_{(1)} = g_{(i)}, i=2,\dots, t$. Get the $p$-values $p_2,\dots,p_t$.

$\bullet$ Analogous to a Bonferroni correction, use $\sum_{i=2}^t p_t$, the upper bound for making at most one Type I error, as the $p$-value of the multiple comparison.

This test aggregates all significance levels into one. The Bonferroni correction will result in a conservative estimate as we ignore much of the correlation structure of the splits. In this scenario, the updates of sample size made in sequential testing should also be adapted as we are now taking the aggregated significance level. A quick and feasible fix is to replace the $p_n$ in (\ref{eqn:ad1}) by the aggregated significance level. Alternatively, we may just test between the best two splits.

Because of the computational cost, when we have two splits that cannot be distinguished, the sequential and multiple testing procedure may end up demanding an extremely large number of points to make the test significant. In practice, we halt the testing early at a cutoff of certain amount $N_{ps}$ of points, and choose the current best split. This compensation for computation time might lower the real power of the test, leading to a less stable result.

\section{Stably Approximating a Black Box} \label{sec:stabletrees}

To build an approximation tree, we replace the greedy splitting criterion by our stabilized version within the CART construction algorithm. At each node, we first generate an initial number of pseudo samples belonging to this node from the black box. Then we compare prospective splits simultaneously based on this set and decide whether we choose the one with the smallest Gini index with certain confidence or request more pseudo sample points. In the latter case, we keep generating until the pseudo sample size reaches what is required by the sequential testing procedure. This is repeated until we distinguish the best split. We perform this procedure on any node that needs to split during construction to get the final approximation tree.

\begin{algorithm}
\DontPrintSemicolon
\KwData{Black box predictor $\mathcal{F}$, covariate distribution of $X$}
\KwResult{Approximation tree $\mathcal{T}$}
\SetKwProg{Fn}{Function}{}{}
\SetKwFunction{SplitNode}{SplitNode}
\Fn{\SplitNode{node V}}{
    \eIf{V satisfies stopping condition}{
        stop and return
    }{
        generate $n$ pseudo samples from $\mathcal{F}$ \;
        find prospective splits $G_1,\cdots, G_m$.\;
        \While{cannot distinguish the best split among $G_1,\cdots, G_m$}{
            generating more pseudo samples whose size is decided by the sequential testing
        }
        claim the best split among $G_1,\cdots, G_m$ and split V by it\;
        \SplitNode{V's left child} \;
        \SplitNode{V's right child}\;
    }
}
$\mathcal{T} \leftarrow$ \SplitNode{root} \;
\caption{Approximation Tree}
\end{algorithm}


In Section \ref{sec:stoppingrules} we explore a potential stopping rule to decide the maximal depth of the tree, although we will often want to impose a maximal depth. We also need to specify $N_{ps}$: the maximal number of pseudo-samples to use and $\alpha$: our risk tolerance for choosing an alternative split.  Additionally we specify which candidate splits to consider, and how to generate the pseudo-sample covariates below.

\subsection{Choice of Prospective Splits}

Most methods of finding prospective splits for a decision tree are compatible with our method once they target  optimizing some information gain \citep{quinlan2014c4, quinlan1987generating}. In building an approximation tree, we only consider making splits at those points which would have been employed in a tree generated from the original training data. We look at the original samples that have been carried along the path and take the possible combinations of the covariates and their middle points of adjacent values that have appeared in those samples. This significantly narrows the number of splits to consider, improving the Bonferroni bounds above, and ensures that the set of candidate splits is independent of the pseudo-sample data.


Although this method will  initially generate a large number of prospective splits, because of the sequential testing scheme, most of those splits will be identified as far worse than the best after a few tests and can be discarded, leaving a negligible effect on the overall performance. In practice, we implement a scheme \citep{benjamini1995controlling} to adaptively discard splits that perform far worse than the current best. All splits are ordered by their p-values against the current best split, and the splits fall below the threshold $\alpha$ are discarded and the Bonferroni procedure applied to the remaining candidates.

\subsection{Generating Points}
\label{Generating Points}
We first generate pseudo covariates and then obtain predictions from the black box to get the responses. It is worth noticing that the first step here may encounter the obstacle that, in practice, we do not have the true generative distribution of covariates.

The methods we develop here do not rely on a particular method of generating pseudo-samples. However, in this paper, we employ a kernel smoother of the empirical covariate distribution in order to maintain a close approximation to the original data (this assumes that these data are available).  This translates to generating pseudo covariates from observed covariates plus random Gaussian noise. In the case of discrete covariates, we choose a neighboring category with a small probability. The width of the kernels to be employed (i.e. the size of perturbation of the training data) may be chosen based on prior knowledge of where new predictions are to be made, or can be chosen by cross-validation.

When we go further down the approximation tree, the covariate space may be narrowed down by the splits along the path. A feasible covariate generator can thus be produced by only smoothing the empirical distribution of those original samples that have been carried on by this path. We further check the boundary condition to ensure that the covariates we generated agree within the region divided by the splits along the path.

\subsection{Performance on Simulated Data}

We use Random Forests (RF) as the teacher model, which will allow us to develop stopping rules below. However, our calculations for developing a tree do not rely on using RF as a teacher and any other machine learning algorithm could be readily employed.

We experiment our method on a simple simulated dataset to check its behavior. Assume $\tilde{X} \in \mathbb{R}^5$ and $\tilde{Y} \in \{0, 1\}$, and let the covariate distribution $\tilde{X} = (x_1,\dots,x_5) \sim \mbox{Unif}[0,1]^5.$ Write $p=P(\tilde{Y}=1|\tilde{X})$ and let
\begin{gather*}
\mbox{logit}(p) =
\begin{cases}
2, &x_1>0.5, \quad x_2>0.7,\\
-3, &x_1>0.5,  \quad 0.7 \geq  x_2>0.2,\\
-4, &x_1>0.5,  \quad x_2 \leq 0.2, \\
3, &x_1 \leq 0.5,  \quad x_5 \leq 0.5, \quad x_3+x_4^2 \geq 1.4, \\
2, &x_1 \leq 0.5,  \quad x_5 \leq 0.5, \quad 1.4 > x_3+x_4^2 \geq 0.5, \\
-2,  &x_1 \leq 0.5,  \quad x_5 \leq 0.5, \quad x_3+x_4^2 < 0.5, \\
2, &x_1 \leq 0.5,  \quad x_5 > 0.5.
\end{cases}
\end{gather*}
The generative distribution is intentionally set to be almost tree-structured so the result should reflect our method working under ideal conditions. We do so to avoid extreme cases during our check, while general distributions will be tested on using real datasets.

We compare across three methods: classification trees (CART), Random Forests (RF) and our proposed approximation tree (AppTree). During each replication, we generate 1,000 sample points from above distribution and obtain a standard RF consisting of 100 trees and a 5-layer CART tree. Then we build a 5-layer approximation tree via the algorithm above. The significant level $\alpha$ for the split test is set to be 0.1, and the maximal number of pseudo samples at each node $N_{ps}$ is set to be $10^4, 10^5$ and $10^6$ respectively. For each $N_{ps}$ we have 100 replications. To assess stability, we use the same setting above but fix one RF as an oracle and learn it by an approximation tree 100 times with $10^4, 10^5$ and $10^6$ respectively.

\begin{figure}[htbp]
    \centering
    \includegraphics[width=6in]{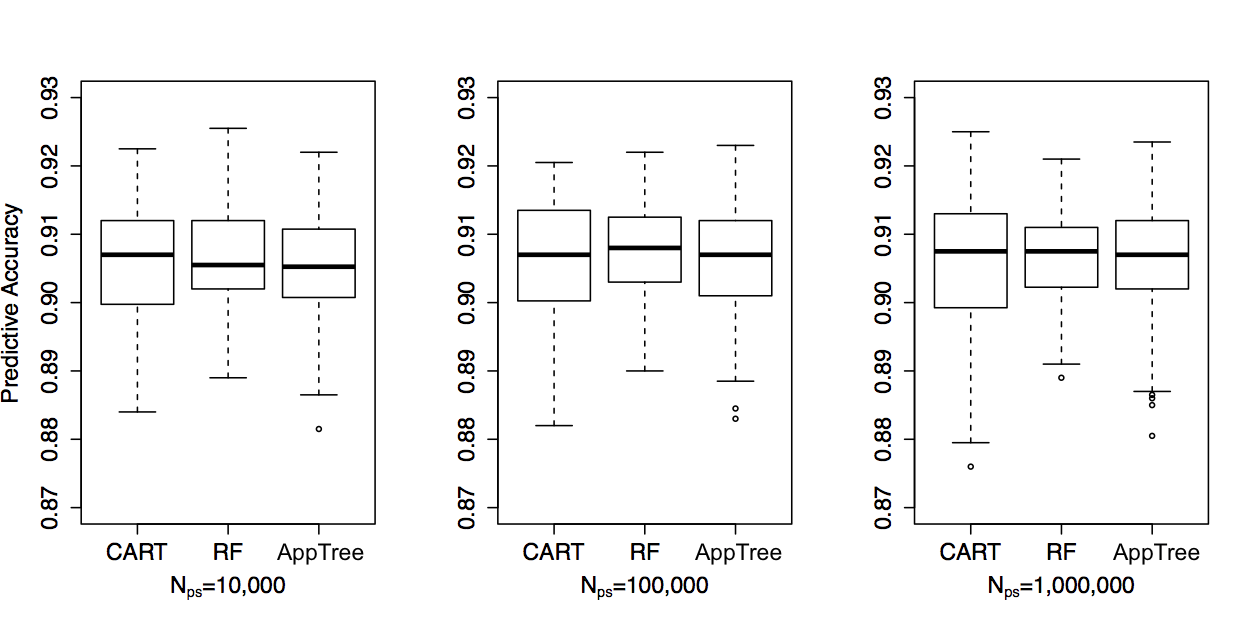}
    \caption{Predictive accuracy of RF, CART and AppTree. Results of RF and CART are recalculated for but are theoretically not affected by different values of $N_{ps}$.}
    \label{fig:PA1}
\end{figure}

\begin{figure}[htbp]
    \centering
    \includegraphics[width=6in]{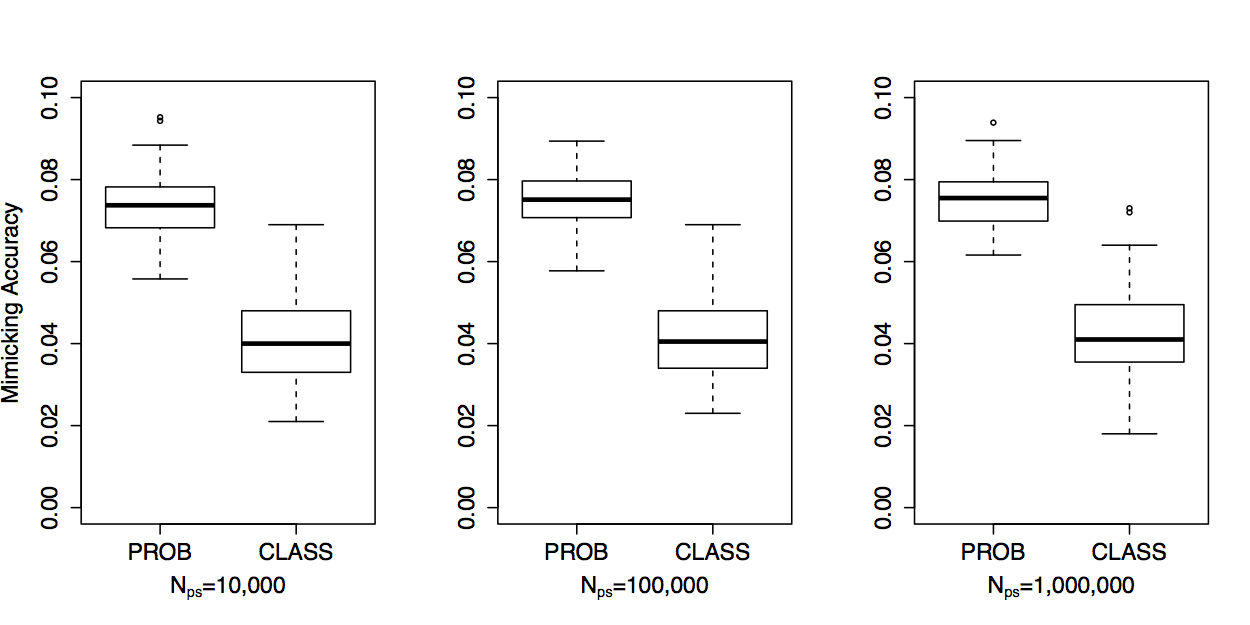}
    \caption{Approximation accuracy. PROB compares RF and AppTree by the $L^1$ difference of their class probabilities. CLASS compares by the predicted class labels.}
    \label{fig:MA1}
\end{figure}

\begin{figure}[htbp]
    \centering
    \includegraphics[width=6in]{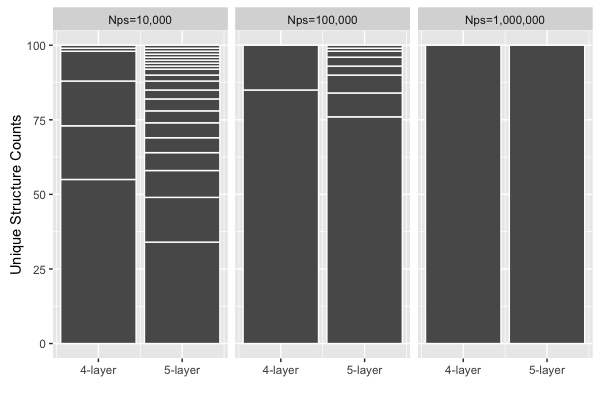}
    \caption{Stability of AppTree with different Nps values. The top 4 layers and top 5 layers of the trees are summarized respectively. In each column, a single black bar represents a unique structure of the tree, while the height of the bar represents the number of occurrence of that structure out of 100 replications.}
    \label{fig:strCnt}
\end{figure}

In order to evaluate predictive accuracy and consistency, we generate new covariates and measure how much the predictions of approximation tree agree with those of the RF. To measure stability, defined in our case as  structural uniqueness, we construct multiple approximation trees out of a single RF and examine the variation in their structures. The split test does not always guarantee a consistent pick through multiple trials due to the pseudo-sample randomness, hence we hope to see small variation among all the trees built. We also examine the trees at different depths to capture the variation along the tree growth. In this paper, our interest lies in the consistency of our approximation trees with RF and the stability of their structures. However, we will still compare the predictive accuracy of approximation trees with other models.

Figure \ref{fig:PA1} shows the predictive accuracy of the three methods on new test points. On average they share similar predictive accuracy; RF has the smallest variance, followed by AppTree. This meets our expectation that AppTree is capable of inheriting stability from the RF after learning how the RF extrapolates. Since the relation between the covariates and the responses is relatively simple, increasing $N_ps$ does not significantly improve the performance.

Figure \ref{fig:MA1} shows the comparison between RF and AppTree in terms of the $L^1$ difference of their predicted class probability, and the disagreement of their class labels. Again the increase of $N_{ps}$ does not bring significant improvement to performance. AppTree has achieved \zycone{95\%} agreement on average with the RF. By expanding the trees to larger sizes the approximation accuracy can still be marginally increased by ``overfitting'' the RF.

Figure \ref{fig:strCnt} shows the stability of AppTree viewing from its top 4 layers and top 5 layers. It can be seen that by increasing the cap on the maximal number of pseudo samples AppTree can generate, its stability is significantly improved. One unique structure is obtained when $N_{ps}=10^6$, which means that some nodes require $\sim 10^6$ points to detect the best split. Two key observations can be made here. Our control of $\alpha$ is relatively conservative due to our sequential testing and multiple testing steps. The maximal number of pseudo sample needed may be quite large to detect the best split. Overall, this initial check shows results as we expected.

In the Supplemental Material we report additional empirical studies on both simulated and real data to show how the performance of approximation tree compares with both decision trees and RFs based on prediction accuracy, consistency with the RF (mimicking accuracy), and stability.

\section{Stopping Rules with Random Forests} \label{sec:stoppingrules}

The methods above ensure that we can construct a tree that will reliably exhibit the same structure across repetitions of the pseudo-data generating mechanism. Here we present a means of deciding how deep the resulting tree can be grown. Generically, this depth may be chosen by the user, or via test-set performance \citep[the approach taken in][]{gibbons2013computerized}.  However, when Random Forests are used as a teacher, recent results on their statistical properties \citep{mentch2016quantifying,wager2017estimation} allow us to determine a stopping rule by asking whether the Random Forests predictions within the current node are statistically different from constant. That is, we ask whether splitting the node further is simply modeling noise.  This allows us to annotate each node with a p-value that tests whether there is further signal within that node.  We view this as an additional criterion for stopping; intelligibility or performance criteria may suggest stopping even with a small p-value -- for example if a class prediction won't change across the node even if the class probability varies -- but that large p-values do indicate that further splits are potentially misleading.  To carry this out, we first review the properties of Random Forests which we will use, and then develop our specific testing criteria.

\subsection{Random Forest Variance}

The tests we propose are based on recent results on the asymptotic normality of random forests. These results consider a model in which the forest is a sum of $m_n$ trees, each built using a subsample of size $k_n$ providing the representation
\[
\hat{F}(x) = \frac{1}{m_n} \sum_{b=1}^{m_n} T_b(x;Z_{S_b(1)},\ldots,Z_{S_b(k_n)},\omega_b)
\]
in which $Z_i = (X_i,Y_i)$ is a row, including response, of the original data, $S_b \subset (1,\ldots,n)$ is a subset of indices selected at random and $T$ is a function that builds a tree using the $Z$'s as data with randomization parameter $\omega$ (used to choose candidate covariates to split on) and makes a prediction for the covariate values $x$. Below we will use the shorthand $Z_{S_b}$ for the collection $Z_{S_b(1)},\ldots,Z_{S_b(K_n)}$. \citet{mentch2016quantifying} observed that this can be seen as an infinite-order incomplete U-statistic with a random kernel whose values are asymptotically normal with variance given by
\begin{equation} \label{eq:rfvar}
\tau^2 = \frac{k_n^2}{n} \zeta_{1,k_n} + \frac{1}{m_n} \zeta_{k_n,k_n}
\end{equation}
with
\[
\zeta_{c, k_n} = \text{cov} (T(Z_1, \ldots, Z_{k_n},\omega), T(Z_1, \ldots, Z_c, Z'_{c+1}, \ldots, Z'_{k_n},\omega'))
\]
the covariance of two trees who share the first $c$ data points. Similar results are found in \citet{wager2017estimation} with differing details on rates and assumptions about $T$. Below we make use of a multivariate extension of these results to predictions at multiple values of $x$ in which the $\zeta_{c,k_n}$ are matrices with dimension corresponding to the number of query points.

Estimates for $\tau^2$ have been proposed in \citet{sexton2009standard,wager2014confidence} as well as in the papers mentioned above; see \citet{athey2016generalized} for demonstrations of consistency of some of these estimates. Here we use query points $x_1,\ldots,x_q$ and employ the infinitesimal jackknife \citep{wager2014confidence} to estimate :
\begin{equation} \label{eq:zeta1}
\left[\hat{\zeta}_{1,k_n}\right]_{kl} = \sum_{i=1}^n  C^i_kC^i_l
\end{equation}
where
\[
C^i_k = \mbox{cov}_b\left(  T_b(x_k;Z_{S_b},\omega_b), 1_{i \in S_b} \right)
\]
is the empirical covariance over the $b = 1,\ldots,B$.
We also use the natural estimate
\begin{align} \label{eq:zetak}
& \left[\hat{\zeta}_{k_n,k_n}\right]_{kl} = \\ & \hspace{0.5cm} \frac{1}{m_n} \sum_{b=1}^{m_n} \left( T_b(x_k;Z_{S_b},\omega_b) - \hat{F}(x) \right)\left( T_b(x_l;Z_{S_b},\omega_b) - \hat{F}(x) \right). \nonumber
\end{align}
See   \citet{ghosal2018boosting} for the consistency of the multivariate infinitesimal jackknife.

Note that while these particular estimates are specific to bagging-type teachers, the procedures below can be readily modified for any other teacher for which a CLT can be proven and where a variance estimate is available, e.g. \citet{ZhouHooker2018}.

\subsection{Testing Procedures}

We apply these results to develop a stopping rule via a test that the variability of $\hat{F}$ within the current node is different from chance. To do so, we employ a variant of the procedures in \citet{mentch2017formal} used to test for structure in Random Forests. Formally, for a node $M$ (which designates a region of feature space), we generate $s$ data points $X_1,\ldots,X_s \in M$ from the kernel density defined above and for which we have the asymptotic result that
\[
(\hat{F}(X_1),\ldots,\hat{F}(X_s)) \sim N( \boldsymbol{\mu}, \Sigma)
\]
and test the hypothesis that $\boldsymbol{\mu} = \mu_{\boldmath{1}}$ is constant over predictions. Formally, $\hat{F}$ has a Gaussian process limit, so that this test can be viewed as an approximation to the hypothesis that $E F(x)$ is constant over $x \in M$.

To carry this out, let $I$ be the $s \times s$ identity matrix and $J$ be the $s \times s$ matrix whose entries are all 1 and we let $\hat{\Sigma}$ be the result of plugging \eqref{eq:zeta1} and \eqref{eq:zetak} into \eqref{eq:rfvar}. Then by standard arguments under the null hypothesis
\[
 \left[ \left(I-\frac{1}{s}J\right)\boldsymbol{\hat{\mu}}\right]^T \left[\left(I-\frac{1}{s}J\right)\hat{\Sigma}\left(I-\frac{1}{s}J\right)\right]^{-1} \left[\left(I-\frac{1}{s}J\right)\boldsymbol{\hat{\mu}}\right] \sim \chi^2_s
\]
from which we can obtain a p-value. This is formalized in Algorithm \ref{alg:testing}.

\begin{algorithm}
\DontPrintSemicolon
\KwData{Information about subsampled Random Forests and the approximation tree; A node in the tree being tested; Number of test points $s$}
\KwResult{p-value at this node}

Randomly generate $s$ test points in S, denoted as $x_1, \ldots, x_s$

Calculate predictions $\boldsymbol{\hat{\mu}} = (F(x_1), \ldots, F(x_s))$ and covariance matrix $\Sigma^{s \times s}$

Calculate test statistics
$$
t =    \left[ \left(I-\frac{1}{s}J\right)\boldsymbol{\hat{\mu}}\right]^T \left[\left(I-\frac{1}{s}J\right)\hat{\Sigma}\left(I-\frac{1}{s}J\right)\right]^{-1} \left[\left(I-\frac{1}{s}J\right)\boldsymbol{\hat{\mu}}\right]
$$

Return the  p-value  $P_{X \sim \chi^2_s} (X > t)$.
\caption{Testing at a given node}\label{alg:testing}
\end{algorithm}

Our choice of $s$ represents a further tuning parameter. Theoretically, larger $s$ is always an improvement, but this must be balanced both with computational cost and with instability in the estimate and inversion of $\hat{\Sigma}$ which can require building very large numbers of trees. \citet{mentch2017formal} employed random projections to improve the stability of these tests for large $s$. Here we take a simpler, though more conservative, approach of generating several sets of $s$ points and averaging the $p$-values; in our studies below we average the $p$-values over 10 sets of $s=10$ query points.

\subsection{Simulation Example}

We demonstrate the behavior of this stopping rule with a simulation example.  Here we suppose our features are two dimensional $(X_1, X_2)$, and $X_i, i = 1, 2$ are each generated independently by a uniform distribution over $[0, 1]$ and we simulate a response from
  \[
    Y=\left\{
    \begin{array}{ll}
    -1,  X_1 < 0.5\\
    +1,   X_1 \geq 0.5\\
    \end{array}
    \right.
  \]
to which we add Gaussian noise with standard deviation 1. Thus, the underlying signal can be represented with a two-level tree.

\begin{figure}[htbp]
    \centering
    \includegraphics[width=6in]{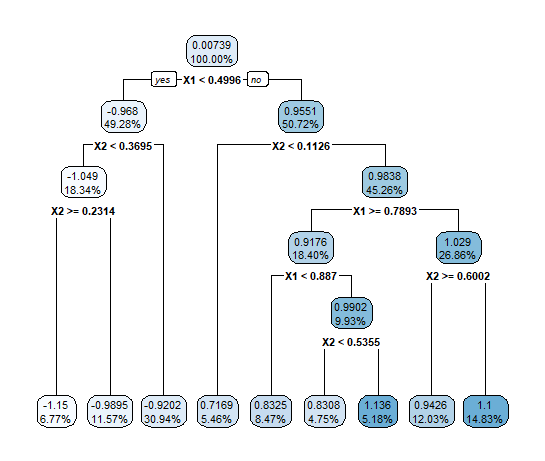}
    \caption{Approximation tree built with \textbf{cp} $= 0.001.$ Dataset is generated by $X_1  \sim U[0, 1]$ and $X_2  \sim U[0, 1]$, independent of each other. $Y + 1 \sim N(0, 1)$ when $X_1 < 0.5$ and $Y - 1 \sim N(0, 1)$ when $X_1 \geq 0.5$. The number of trees $B = 10,000$.}
    \label{fig:S3}
\end{figure}

We generated 500 data sets of size 1,000 from this process and trained a Random Forests of 10,000 trees on each using 200 observations for each tree. The large number of trees was chosen to stabilize our variance estimate. We then used 10,000 test points to generate an approximation tree, and example of which is given in Figure  \ref{fig:S3}.

%

\begin{figure}[htbp]
    \centering
    \includegraphics[width=6in]{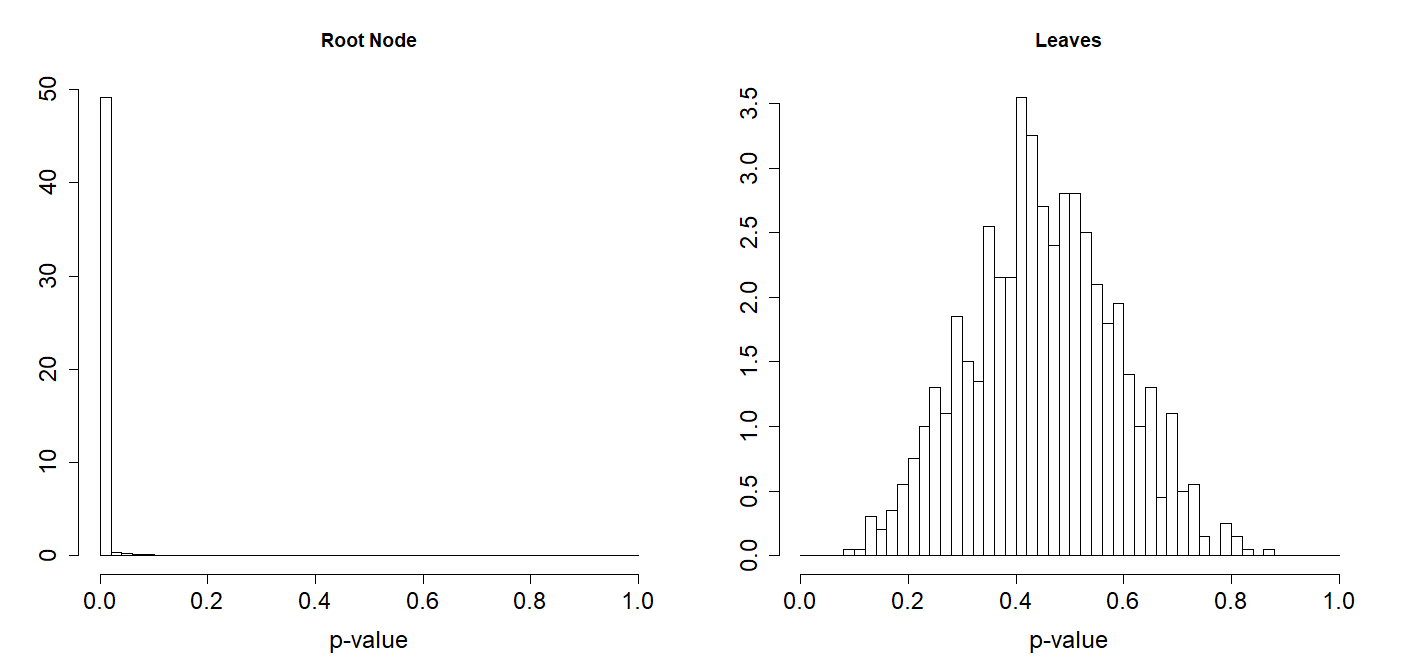}
    \caption{Histograms for p-values based of 500 simulated data set with a three-node tree as an underlying generator. Left plot gives $p$-values at the root node, with power 0.994 to detect a real split. Right gives $p$-values at the children where no further splitting is warranted. The non-uniform distribution of the right-hand plot is due to averaging $p$-values over 10 sets of 10 query points.}
    \label{fig:hist}
\end{figure}

For these data, the approximation trees all split on $X_1$ at the root note. Figure \ref{fig:hist} records the $p$-values we calculate at the root node over all 500 data sets, and the combined p-values of the left and right children (since symmetry arguments give these the same distribution).  Using a 0.05 threshold, all but 3 of the approximation trees would be chosen to be split at the root node and no trees exhibit further splits.

This gives us confidence that we can distinguish appropriate tree structure when it exists. However, as we observe in our real-world examples below, the underlying structure may not be well approximated by a shallow tree and we may wish to truncate an approximation tree, even if there is remaining signal that can be distinguished from noise.

%


\section{Real-world Experiments} \label{sec:examples}

Here we present two case-studies in using our approximation tree algorithm. In each of these we first obtain a subsampled Random Forests to predict the response using the \texttt{randomForest} package in R.  We then develop a stabilized approximation tree for that forest using up to 500,000 pseudo-samples per split. We also annotate each node with a p-value indicating whether the values of the underlying Random Forests are statistically distinguishable from being constant. For our case studies, we find that we stop due to the maximum tree depth being reached before the observed p-values become large.

\subsection{CAD-MDD Data}

The CAD-MDD data were employed in \citet{gibbons2013computerized} to shorten diagnostic screening questionnaires for severe depression. The data set contains 836 responses to an 88-question tool along with a clinical diagnosis. The length of this tool represents a significant response burden and a goal is to shorten the tool while retaining diagnostic accuracy by adaptively asking questions as the subject answers them.

\begin{figure}[ht]
   \centering
   \includegraphics[width=0.8\linewidth]{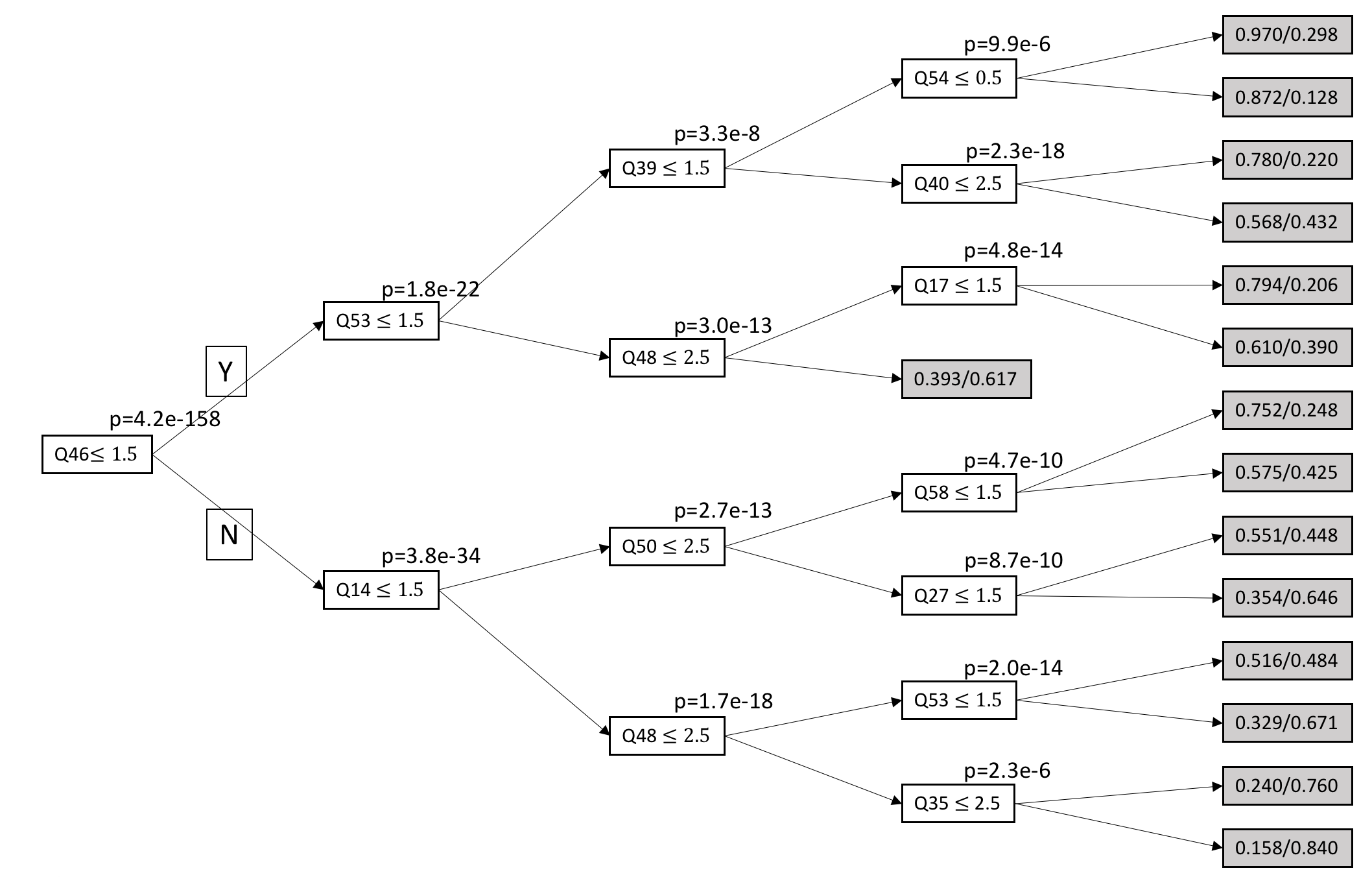} 
   \caption{Adaptive screening tool extracted from the dominant tree structure for CAD-MDD data.}
   \label{fig:mddtree}
\end{figure}

Figure \ref{fig:mddtree} presents the dominant tree structure, occurring in 86 out of 100 replications that we obtain out of the approximation tree algorithm, annotated with significance levels for each split.   The pseudo-code-formatted decision rule can be followed by adaptively asking patients one question at a time until arriving at a leaf node. Following \citet{gibbons2013computerized} we improve the depression diagnosis from an 88-question survey plus a RF predictor, to a unique adaptive screening tool with at most 5 questions, while retaining 90\% consistency. It is now possible for psychiatrists to manually follow the screening tool asking a few questions to perform a quick initial diagnosis for patients without directly going through the 88-question survey, or adaptive data collection can also be automated.

In order to demonstrate that each split we made is indeed capturing true signal instead of random variability, each intermediate tree node is annotated with a p-value returned by Algorithm \ref{alg:testing}. Full details of the learned approximation tree can be found in the Supplemental Material.

We also note that although we only have 836 original patients, we may still need $5\times 10^5$ pseudo points to  stabilize a split. One possible reason is that we have many variables and values to choose as a splitting rule, making the best splits indistinguishable and always reaching the pseudo sample size cap $N_{ps}$. An alternative remedy is to obtain a prior set of fewer splits of interests and split by this set. However, in general we still require a large number of points.

\subsection{COMPAS Data}

COMPAS, abbreviated for Correctional Offender Management Profiling for Alternative Sanctions, is a proprietary score developed to predict recidivism risk. It has been the subject of considerable controversy, stimulating a growing literature on the subject of bias in algorithmic decision making; see \citet{angwin2016machine}, \citet{kleinberg2016inherent} or \citet{tan2017detecting} for example. ProPublica analyzed the COMPAS Recidivism Algorithm \citep{larson2016we} and released the dataset containing criminal history, jail and prison time, demographics and COMPAS scores for defendants from Broward County, Florida. \footnote{https://github.com/propublica/compas-analysis}

\begin{figure}[ht]
   \centering
   \includegraphics[width=0.8\linewidth]{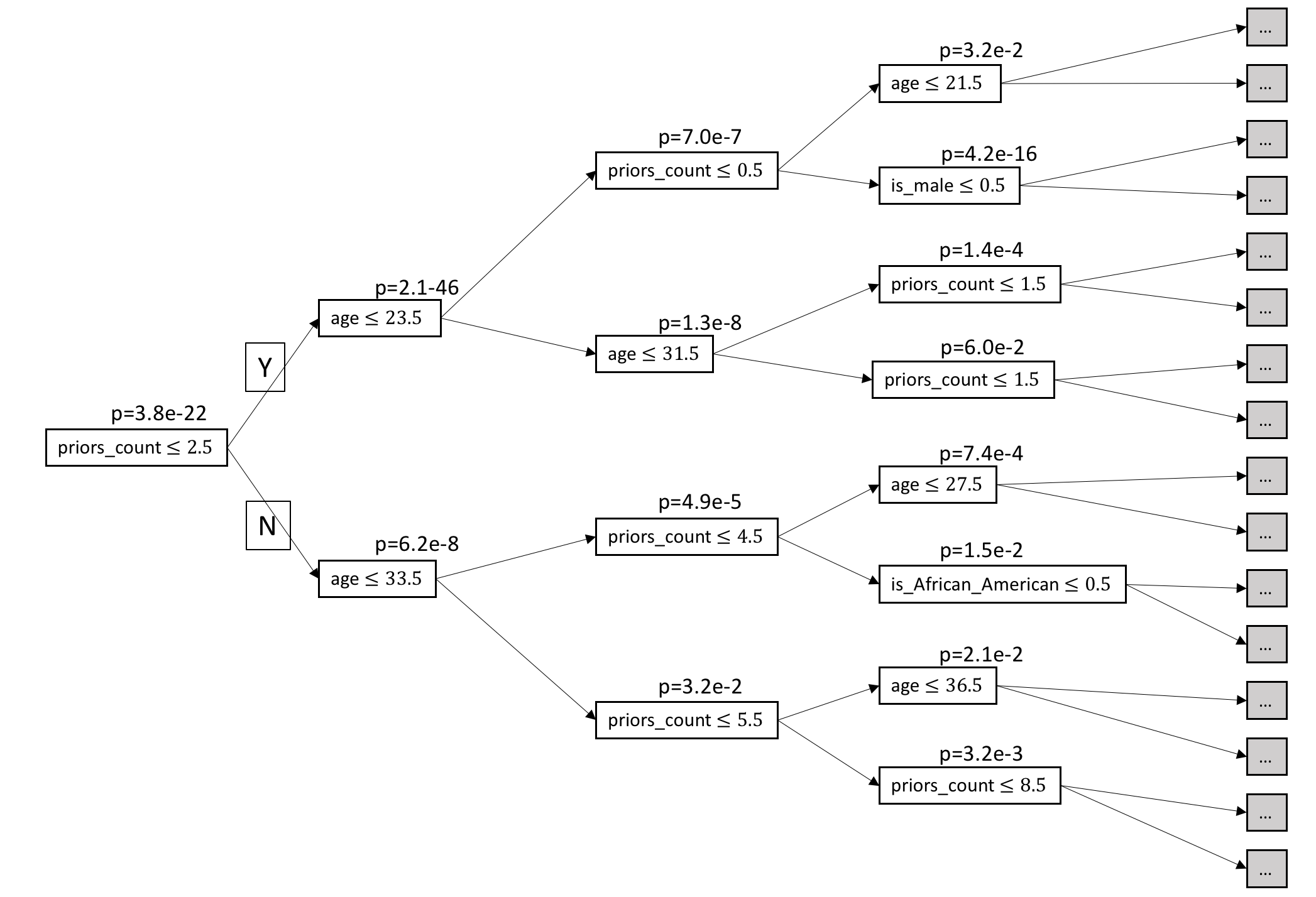} 
   \caption{Compas Tree after truncation.}
   \label{fig:compastree}
\end{figure}

Our focus here is to build a stable and reliable approximation tree, which may then be investigated for explicit racial bias. We select nine features and reformulate the task as a binary classification problem: recidivism or not. Due to space constraints, Figure \ref{fig:compastree} shows part of our approximation tree with p-values annotated; see Supplemental Material for details.

%
\section{Conclusion}

While much attention has been given to ways to understand the result of machine learning algorithms or to provide explanations for particular predictions, there has been relatively little attention given to the statistical stability of those explanations. An explanation for a prediction that might differ due to random chance {\em even when the underlying prediction doesn't change} may be reasonably regarded as spurious.  In this paper, we propose a method to ensure the stability of explanations that are given by decision trees that result from model distillation. We have shown that a hypothesis testing framework can be employed to ensure that sufficient pseudo-data is generated to consistently choose the same tree structure.  Moreover, when distilling a Random Forests we have used recent results on their statistical properties to design stopping rules to ensure that explanations derived from approximating decision trees reflect consistent signal rather than sample noise.

Our empirical results suggest that while trees can be used as students to recover predictions with high accuracy, stabilizing their structure, and hence providing stable explanations for predictions, can require the generation of much larger pseudo-data sets than is currently common. We have produced evidence that our stopping rules will indeed choose the ``right'' tree structure, but we observe that in at least two cases trees of more than five levels continue to uncover real signal.

The framework we have employed here has been based on the use of a Gini split criterion and Random Forests. The first of these can readily be replaced with any criterion that can be represented as an average (subject to mild regularity conditions). Our use of Random Forests is motivated by a central limit theorem for the values of Random Forests predictions; the use of these stopping rules with other machine learning teachers requires similar results for them, of which we know few.

It is important to note that split-stabilization and stopping rules are based on two different sources of variability. We control pseudo-data variability in the choice of splits, while variability from the original sample is calculated for our stopping rule. These choices have been made in the interests of presenting a single tree that retains reasonable fidelity to the teacher. We have not assessed whether sample variability might produce a teacher which induces a different approximation tree; this would require accounting for the covariance of predictions within our central limit theorem for the split criterion as well as intelligible means of representing variability in tree structures and we leave this for future work.

\section*{Acknowledgements}
This work was partially supported by grants NIH R03DA036683, NSF DMS-1053252 and NSF DEB-1353039.
The authors would like to thank Robert Gibbons for providing the CAD-MDD data.

\bibliographystyle{chicago}
\bibliography{citation}

\begin{thebibliography}{}

\bibitem[\protect\citeauthoryear{Angwin, Larson, Mattu, and Kirchner}{Angwin
  et~al.}{2016}]{angwin2016machine}
Angwin, J., J.~Larson, S.~Mattu, and L.~Kirchner (2016).
\newblock Machine bias: There’s software used across the country to predict
  future criminals. and it’s biased against blacks.
\newblock {\em ProPublica, May\/}~{\em 23}.

\bibitem[\protect\citeauthoryear{Athey, Tibshirani, and Wager}{Athey
  et~al.}{2016}]{athey2016generalized}
Athey, S., J.~Tibshirani, and S.~Wager (2016).
\newblock Generalized random forests.
\newblock {\em arXiv preprint arXiv:1610.01271\/}.

\bibitem[\protect\citeauthoryear{Augasta and Kathirvalavakumar}{Augasta and
  Kathirvalavakumar}{2012}]{augasta2012reverse}
Augasta, M.~G. and T.~Kathirvalavakumar (2012).
\newblock Reverse engineering the neural networks for rule extraction in
  classification problems.
\newblock {\em Neural processing letters\/}~{\em 35\/}(2), 131--150.

\bibitem[\protect\citeauthoryear{Banerjee, McKeague, et~al.}{Banerjee
  et~al.}{2007}]{banerjee2007confidence}
Banerjee, M., I.~W. McKeague, et~al. (2007).
\newblock Confidence sets for split points in decision trees.
\newblock {\em The Annals of Statistics\/}~{\em 35\/}(2), 543--574.

\bibitem[\protect\citeauthoryear{Benjamini and Hochberg}{Benjamini and
  Hochberg}{1995}]{benjamini1995controlling}
Benjamini, Y. and Y.~Hochberg (1995).
\newblock Controlling the false discovery rate: a practical and powerful
  approach to multiple testing.
\newblock {\em Journal of the Royal Statistical Society. Series B
  (Methodological)\/}, 289--300.

\bibitem[\protect\citeauthoryear{Breiman}{Breiman}{2001}]{breiman2001random}
Breiman, L. (2001).
\newblock Random forests.
\newblock {\em Machine learning\/}~{\em 45\/}(1), 5--32.

\bibitem[\protect\citeauthoryear{Breiman, Friedman, Stone, and Olshen}{Breiman
  et~al.}{1984}]{breiman1984classification}
Breiman, L., J.~Friedman, C.~J. Stone, and R.~A. Olshen (1984).
\newblock {\em Classification and regression trees}.
\newblock CRC press.

\bibitem[\protect\citeauthoryear{Chouldechova}{Chouldechova}{2017}]{chouldechova2017fair}
Chouldechova, A. (2017).
\newblock Fair prediction with disparate impact: A study of bias in recidivism
  prediction instruments.
\newblock {\em Big data\/}~{\em 5\/}(2), 153--163.

\bibitem[\protect\citeauthoryear{Craven and Shavlik}{Craven and
  Shavlik}{1995}]{craven1995extracting}
Craven, M.~W. and J.~W. Shavlik (1995).
\newblock Extracting tree-structured representations of trained networks.
\newblock In {\em NIPS}.

\bibitem[\protect\citeauthoryear{Dunnett}{Dunnett}{1955}]{dunnett1955multiple}
Dunnett, C.~W. (1955).
\newblock A multiple comparison procedure for comparing several treatments with
  a control.
\newblock {\em Journal of the American Statistical Association\/}~{\em
  50\/}(272), 1096--1121.

\bibitem[\protect\citeauthoryear{Friedman}{Friedman}{2001}]{friedman2001greedy}
Friedman, J.~H. (2001).
\newblock Greedy function approximation: a gradient boosting machine.
\newblock {\em Annals of statistics\/}, 1189--1232.

\bibitem[\protect\citeauthoryear{Ghosal and Hooker}{Ghosal and
  Hooker}{2018}]{ghosal2018boosting}
Ghosal, I. and G.~Hooker (2018).
\newblock Boosting random forests to reduce bias; one-step boosted forest and
  its variance estimate.
\newblock {\em arXiv preprint arXiv:1803.08000\/}.

\bibitem[\protect\citeauthoryear{Gibbons, Hooker, Finkelman, Weiss, Pilkonis,
  Frank, Moore, and Kupfer}{Gibbons et~al.}{2013}]{gibbons2013computerized}
Gibbons, R.~D., G.~Hooker, M.~D. Finkelman, D.~J. Weiss, P.~A. Pilkonis,
  E.~Frank, T.~Moore, and D.~J. Kupfer (2013).
\newblock The computerized adaptive diagnostic test for major depressive
  disorder (cad-mdd): a screening tool for depression.
\newblock {\em The Journal of clinical psychiatry\/}~{\em 74\/}(7), 1--478.

\bibitem[\protect\citeauthoryear{Goldstein, Kapelner, Bleich, and
  Pitkin}{Goldstein et~al.}{2013}]{Goldstein2013}
Goldstein, A., A.~Kapelner, J.~Bleich, and E.~Pitkin (2013, 09).
\newblock Peeking inside the black box: Visualizing statistical learning with
  plots of individual conditional expectation.
\newblock ~{\em 24}.

\bibitem[\protect\citeauthoryear{He, Eisner, and Daume}{He
  et~al.}{2012}]{he2012imitation}
He, H., J.~Eisner, and H.~Daume (2012).
\newblock Imitation learning by coaching.
\newblock In {\em Advances in Neural Information Processing Systems}, pp.\
  3149--3157.

\bibitem[\protect\citeauthoryear{Hooker}{Hooker}{2007}]{hooker2007generalized}
Hooker, G. (2007).
\newblock Generalized functional anova diagnostics for high-dimensional
  functions of dependent variables.
\newblock {\em Journal of Computational and Graphical Statistics\/}~{\em
  16\/}(3), 709--732.

\bibitem[\protect\citeauthoryear{{Hu}, {Chen}, {Nair}, and {Sudjianto}}{{Hu}
  et~al.}{2018}]{Hu2018}
{Hu}, L., J.~{Chen}, V.~N. {Nair}, and A.~{Sudjianto} (2018, June).
\newblock {Locally Interpretable Models and Effects based on Supervised
  Partitioning (LIME-SUP)}.
\newblock {\em ArXiv e-prints\/}.

\bibitem[\protect\citeauthoryear{Johansson and Niklasson}{Johansson and
  Niklasson}{2009}]{johansson2009evolving}
Johansson, U. and L.~Niklasson (2009).
\newblock Evolving decision trees using oracle guides.
\newblock In {\em Computational Intelligence and Data Mining, 2009. CIDM'09.
  IEEE Symposium on}, pp.\  238--244. IEEE.

\bibitem[\protect\citeauthoryear{Johansson, S{\"o}nstr{\"o}d, and
  L{\"o}fstr{\"o}m}{Johansson et~al.}{2010}]{johansson2010oracle}
Johansson, U., C.~S{\"o}nstr{\"o}d, and T.~L{\"o}fstr{\"o}m (2010).
\newblock Oracle coached decision trees and lists.
\newblock In {\em International Symposium on Intelligent Data Analysis}, pp.\
  67--78. Springer.

\bibitem[\protect\citeauthoryear{Johansson, S{\"o}nstr{\"o}d, and
  L{\"o}fstr{\"o}m}{Johansson et~al.}{2011}]{johansson2011one}
Johansson, U., C.~S{\"o}nstr{\"o}d, and T.~L{\"o}fstr{\"o}m (2011).
\newblock One tree to explain them all.
\newblock In {\em Evolutionary Computation (CEC), 2011 IEEE Congress on}, pp.\
  1444--1451. IEEE.

\bibitem[\protect\citeauthoryear{Kleinberg, Mullainathan, and
  Raghavan}{Kleinberg et~al.}{2016}]{kleinberg2016inherent}
Kleinberg, J., S.~Mullainathan, and M.~Raghavan (2016).
\newblock Inherent trade-offs in the fair determination of risk scores.
\newblock {\em arXiv preprint arXiv:1609.05807\/}.

\bibitem[\protect\citeauthoryear{Larson, Mattu, Kirchner, and Angwin}{Larson
  et~al.}{2016}]{larson2016we}
Larson, J., S.~Mattu, L.~Kirchner, and J.~Angwin (2016).
\newblock How we analyzed the compas recidivism algorithm.
\newblock {\em ProPublica (5 2016)\/}~{\em 9}.

\bibitem[\protect\citeauthoryear{Lichman}{Lichman}{2013}]{Lichman:2013}
Lichman, M. (2013).
\newblock {UCI} machine learning repository.

\bibitem[\protect\citeauthoryear{Lou, Caruana, and Gehrke}{Lou
  et~al.}{2012}]{lou2012}
Lou, Y., R.~Caruana, and J.~Gehrke (2012).
\newblock Intelligible models for classification and regression.
\newblock In {\em Proceedings of the 18th ACM SIGKDD International Conference
  on Knowledge Discovery and Data Mining}, KDD '12, New York, NY, USA. ACM.

\bibitem[\protect\citeauthoryear{Lucas, Klein, Tannahill, Ivanova, Brandon,
  Domyancic, and Zhang}{Lucas et~al.}{2013}]{lucas2013failure}
Lucas, D., R.~Klein, J.~Tannahill, D.~Ivanova, S.~Brandon, D.~Domyancic, and
  Y.~Zhang (2013).
\newblock Failure analysis of parameter-induced simulation crashes in climate
  models.
\newblock {\em Geoscientific Model Development\/}~{\em 6\/}(4), 1157--1171.

\bibitem[\protect\citeauthoryear{Mangasarian, Street, and Wolberg}{Mangasarian
  et~al.}{1995}]{mangasarian1995breast}
Mangasarian, O.~L., W.~N. Street, and W.~H. Wolberg (1995).
\newblock Breast cancer diagnosis and prognosis via linear programming.
\newblock {\em Operations Research\/}~{\em 43\/}(4), 570--577.

\bibitem[\protect\citeauthoryear{Mentch and Hooker}{Mentch and
  Hooker}{2016}]{mentch2016quantifying}
Mentch, L. and G.~Hooker (2016).
\newblock Quantifying uncertainty in random forests via confidence intervals
  and hypothesis tests.
\newblock {\em The Journal of Machine Learning Research\/}~{\em 17\/}(1),
  841--881.

\bibitem[\protect\citeauthoryear{Mentch and Hooker}{Mentch and
  Hooker}{2017}]{mentch2017formal}
Mentch, L. and G.~Hooker (2017).
\newblock Formal hypothesis tests for additive structure in random forests.
\newblock {\em Journal of Computational and Graphical Statistics\/}~{\em
  26\/}(3), 589--597.

\bibitem[\protect\citeauthoryear{Quinlan}{Quinlan}{1987}]{quinlan1987generating}
Quinlan, J.~R. (1987).
\newblock Generating production rules from decision trees.
\newblock In {\em IJCAI}, Volume~87, pp.\  304--307. Citeseer.

\bibitem[\protect\citeauthoryear{Quinlan}{Quinlan}{2014}]{quinlan2014c4}
Quinlan, J.~R. (2014).
\newblock {\em C4. 5: programs for machine learning}.
\newblock Elsevier.

\bibitem[\protect\citeauthoryear{Ribeiro, Singh, and Guestrin}{Ribeiro
  et~al.}{2016}]{ribeiro2016should}
Ribeiro, M.~T., S.~Singh, and C.~Guestrin (2016).
\newblock Why should i trust you?: Explaining the predictions of any
  classifier.
\newblock In {\em Proceedings of the 22nd ACM SIGKDD International Conference
  on Knowledge Discovery and Data Mining}, pp.\  1135--1144. ACM.

\bibitem[\protect\citeauthoryear{Sexton and Laake}{Sexton and
  Laake}{2009}]{sexton2009standard}
Sexton, J. and P.~Laake (2009).
\newblock Standard errors for bagged and random forest estimators.
\newblock {\em Computational Statistics \& Data Analysis\/}~{\em 53\/}(3),
  801--811.

\bibitem[\protect\citeauthoryear{Simonyan, Vedaldi, and Zisserman}{Simonyan
  et~al.}{2013}]{simonyan2013deep}
Simonyan, K., A.~Vedaldi, and A.~Zisserman (2013).
\newblock Deep inside convolutional networks: Visualising image classification
  models and saliency maps.
\newblock {\em arXiv preprint arXiv:1312.6034\/}.

\bibitem[\protect\citeauthoryear{Tan, Caruana, Hooker, and Lou}{Tan
  et~al.}{2017}]{tan2017detecting}
Tan, S., R.~Caruana, G.~Hooker, and Y.~Lou (2017).
\newblock Detecting bias in black-box models using transparent model
  distillation.
\newblock {\em arXiv preprint arXiv:1710.06169\/}.

\bibitem[\protect\citeauthoryear{Wager and Athey}{Wager and
  Athey}{2017}]{wager2017estimation}
Wager, S. and S.~Athey (2017).
\newblock Estimation and inference of heterogeneous treatment effects using
  random forests.
\newblock {\em Journal of the American Statistical
  Association\/}~(just-accepted).

\bibitem[\protect\citeauthoryear{Wager, Hastie, and Efron}{Wager
  et~al.}{2014}]{wager2014confidence}
Wager, S., T.~Hastie, and B.~Efron (2014).
\newblock Confidence intervals for random forests: The jackknife and the
  infinitesimal jackknife.
\newblock {\em The Journal of Machine Learning Research\/}~{\em 15\/}(1),
  1625--1651.

\bibitem[\protect\citeauthoryear{{Zhou} and {Hooker}}{{Zhou} and
  {Hooker}}{2018}]{ZhouHooker2018}
{Zhou}, Y. and G.~{Hooker} (2018).
\newblock {Boulevard: Regularized Stochastic Gradient Boosted Trees and Their
  Limiting Distribution}.
\newblock {\em ArXiv e-prints\/}.

\end{thebibliography}

\appendix

\section{Additional results on AppTree} \label{sec:appendix}

\subsection{Real Datasets}

In this section we will show the results of our method on eight datasets. Seven of them are available on the UCI repository \citep{Lichman:2013} and one is the CAD-MDD data used by \citet{gibbons2013computerized}. We manually split each dataset into train and test for cross validation. Table \ref{tab:datasetdes} shows the number of covariates, training and testing sample size and the levels of responses for each dataset.

\begin{table}[htbp]
   \centering
   \begin{tabular}{ccccc} 
      \toprule
      Name & \#Cov & \#Train & \#Test & Response Levels\\
      \midrule
      CAD-MDD & 88 & 500 & 336 & 0,1\\
      BreastCancer\citep{mangasarian1995breast} & 30 & 350 & 218 & 0,1\\
      Car & 6 & 1000 & 727 & 0,1\\
      ClimateModel\citep{lucas2013failure} & 18 & 400 & 140 & 0,1\\
      Abalone & 10 & 3133 & 1044 & 0,1,2,3 \\
      Cardiotocography & 30 & 1126 & 1000 & 0,1,2\\
      WineRed & 11 & 1100 & 499 & 0,1,2\\
      WineWhite & 11 & 3000 & 1898 & 0,1,2\\
      \bottomrule
   \end{tabular}
   \caption{Dataset description showing the number of covariates, the number of training points, the number of testing points and the levels of responses for each dataset.}
   \label{tab:datasetdes}
\end{table}

To decide the generative distribution of covariates before running our algorithm, we perturb the empirical distribution  by Gaussian noise whose variances are approximately 1/50 of the ranges of corresponding covariates. The probability of jumping to a neighboring category for discrete covariates is set to be 1/7.

We compare across four methods here: classification trees (CART), random forests (RF), our proposed approximation tree (AppTree), and a baseline method (BASE). Previous work \citep{johansson2009evolving, johansson2010oracle} fixes the number of pseudo samples from the oracle during the coaching procedure. Analogously, we set BASE to be a non-adaptive version of our AppTree which requests a pseudo sample set from the RF only once at the root node and uses it all the way down. The pseudo sample size is set to be 9 times the size of the training data; larger than the samples used in \citep{johansson2009evolving, johansson2010oracle} and is designed as a reasonable blind decision without any prior information.

We use the same setting for all datasets for consistency. For each dataset, we train a RF containing 200 trees, a CART tree, then 100 AppTrees and 100 BASE trees  approximating the RF. $N_{ps} = 500,000$, which means each node of AppTree can generate at most $5\times 10^5$ pseudo samples to decide its split. CART, BASE and AppTree all grow to the 6th layer including the root. Confidence level $\alpha$ is set to be 0.1.

\subsection{Binary Classification}
\begin{figure}[htbp] 
   \centering
   \includegraphics[width=6in]{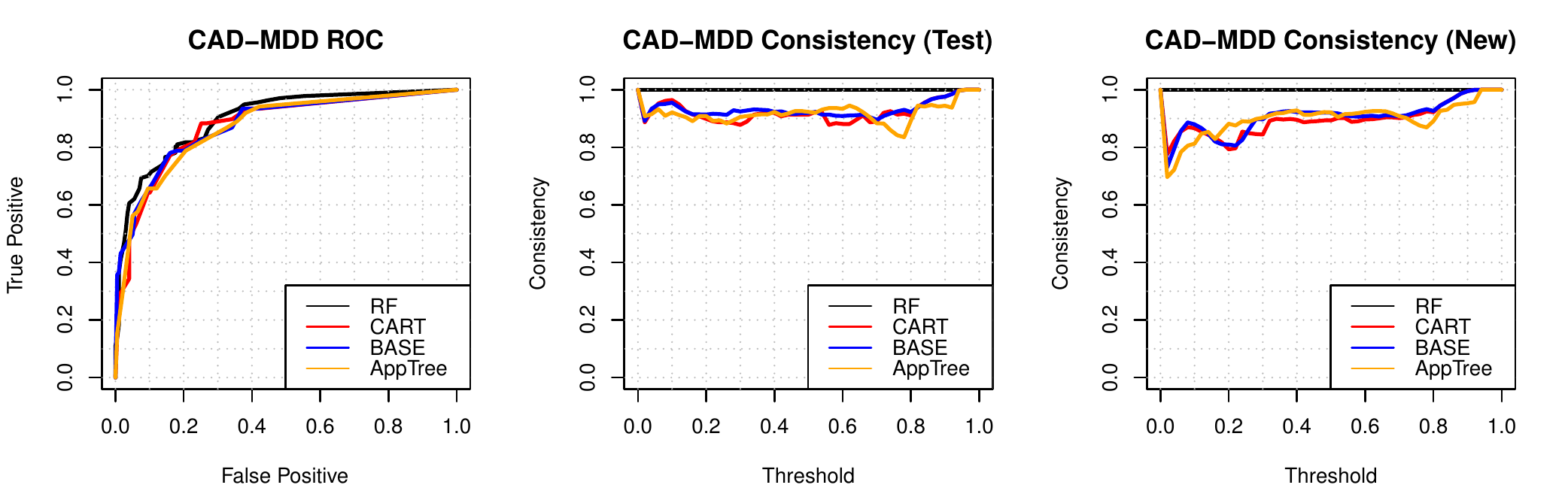}
   \includegraphics[width=6in]{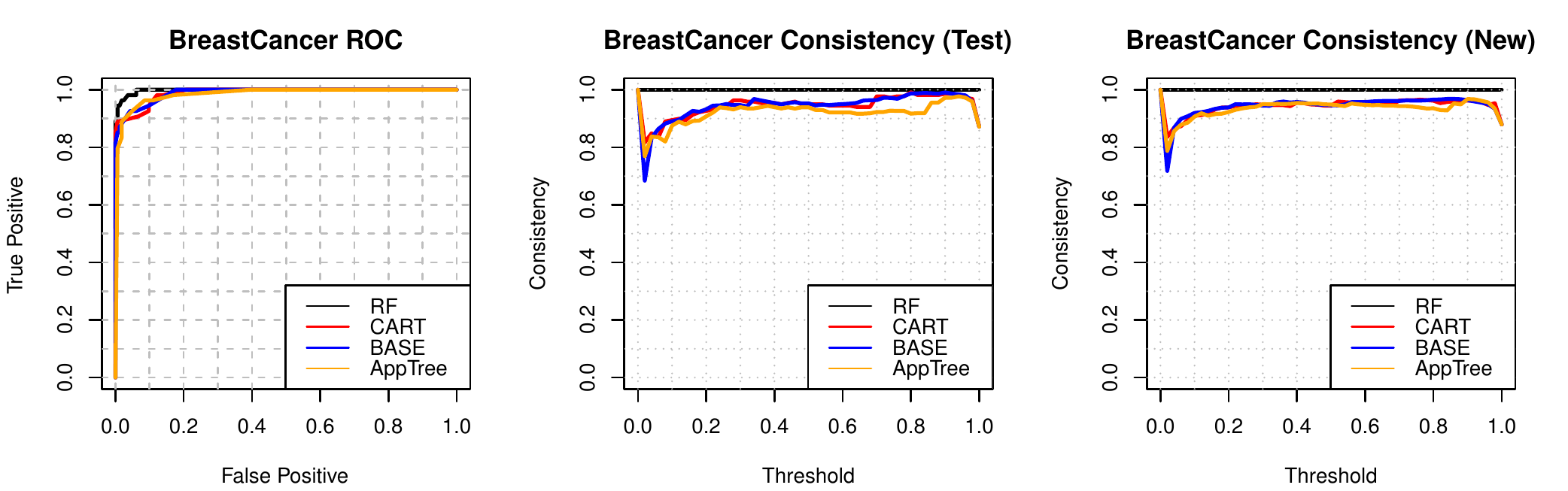}
   \includegraphics[width=6in]{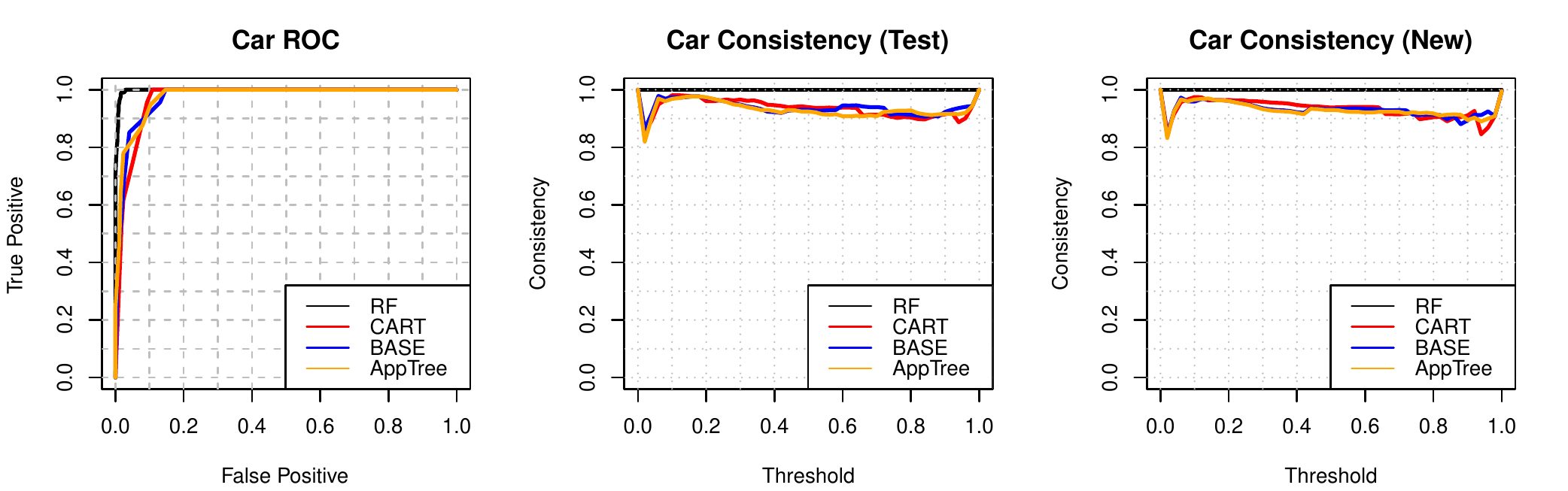}
   \includegraphics[width=6in]{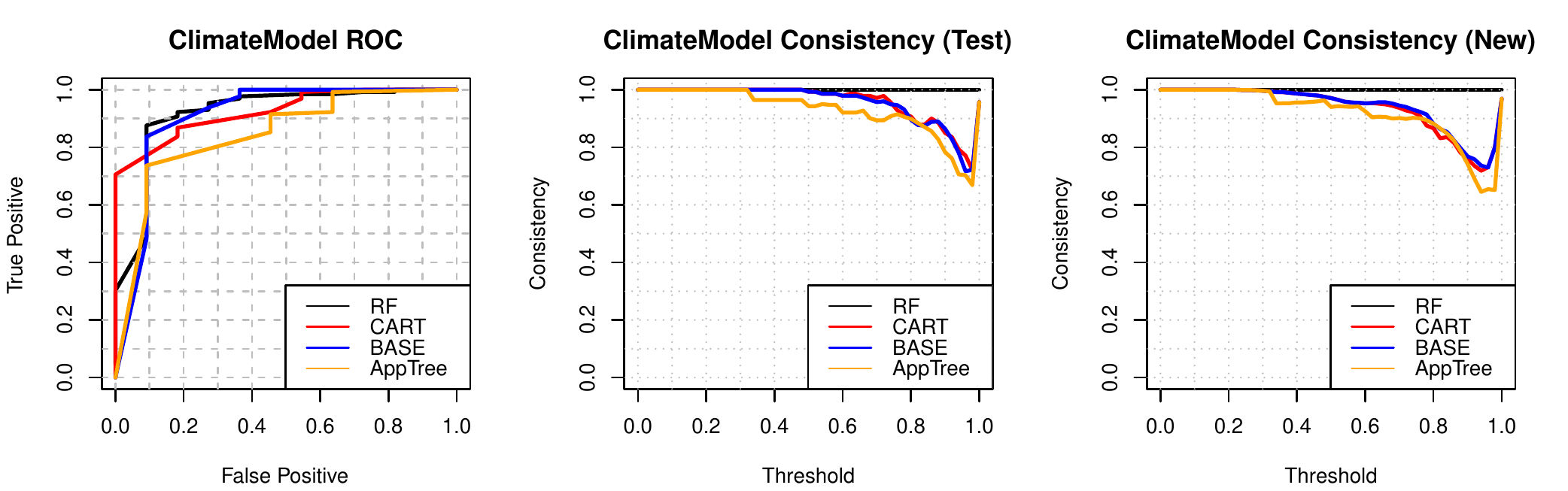}
   \caption{Performance evaluation on binary classification datasets. From top to bottom: CAD-MDD, BreastCancer, Car, ClimateModel. From left to right: ROC curves, consistency with RF on testing set, consistency with RF on new data points.}
   \label{fig:bi1}
\end{figure}

Figure \ref{fig:bi1} shows the evaluation of methods on binary classification datasets. \citet{johansson2011one} has pointed out that a single decision tree is already capable of mimicking an oracle predictor to make highly accurate predictions given the oracle is not overly complicated. Our simulation shows similar results, as all three methods CART, BASE and AppTree tightly follow the ROC curves of the RF and there is no significance difference among them.  Consistency is measured as the frequency of a model agreeing with the RF when predicting on same input covariates. We use a moving threshold as the classification boundary, and evaluate the consistency on both the testing data and the  new data (as marked ``test" and ``new" in the plot). First, all three 6-layer trees can approximate the RF with over 80\% probability for almost any given classification threshold, and there is no significant difference among them. Further, the behaviors of both ``test'' and ``new'' plot seem similar, which lends support to our generative covariate distribution. While the overall 80\% consistency may seem not powerful enough to make those trees aligned with the oracle RF, we can build the trees deeper to better ``overfit'' the RF.

\subsection{Multiclass Classification}

\begin{figure}[htbp] 
   \centering
   \includegraphics[width=6in]{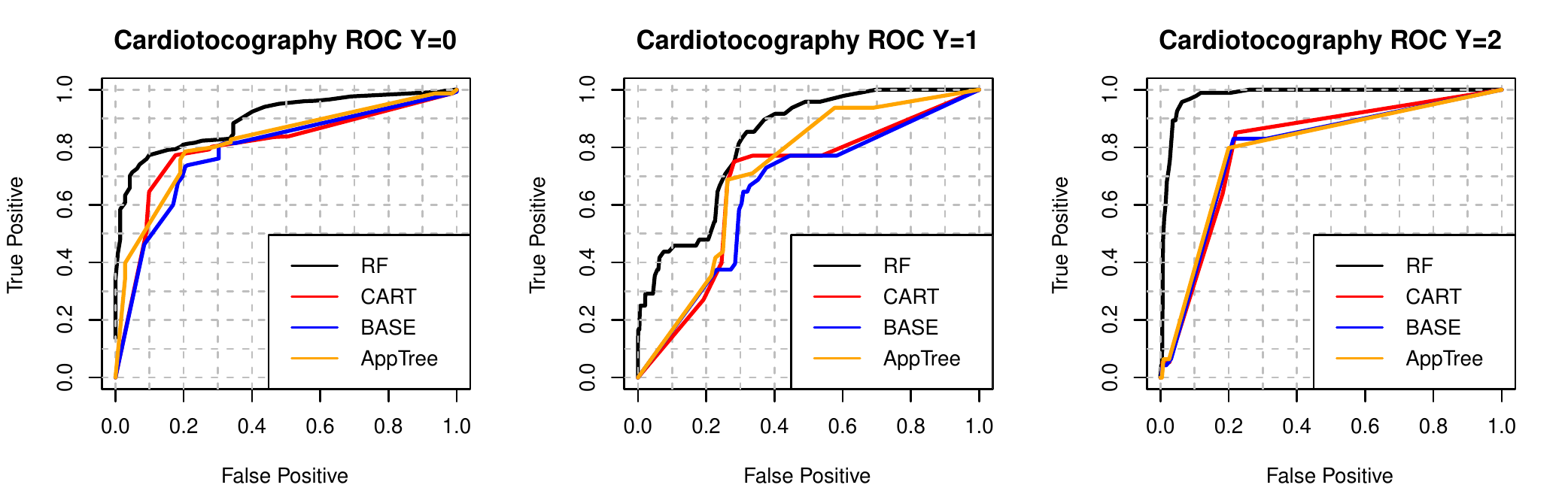}
   \includegraphics[width=6in]{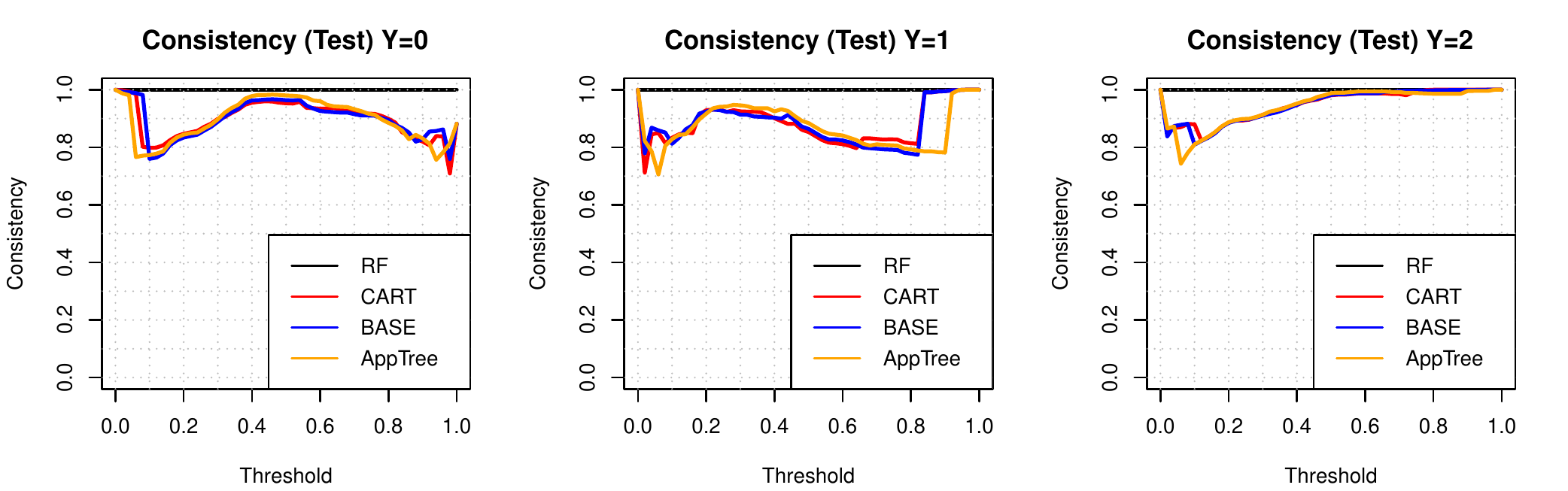}
   \caption{Performance evaluation on multiclass (3-class) classification dataset Cardiotocography. ROC curves are plotted in a one v.s. all fashion. Consistency is only checked on testing data.}
   \label{fig:mi11}
\end{figure}

\begin{figure}[htbp] 
   \centering
   \includegraphics[width=6in]{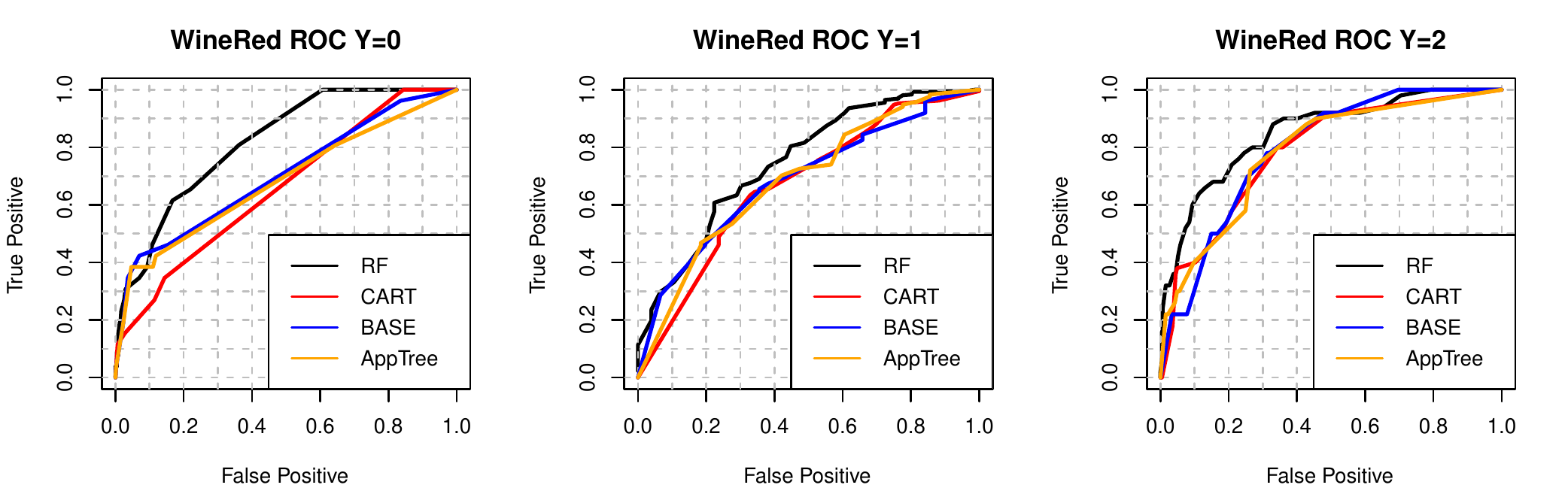}
   \includegraphics[width=6in]{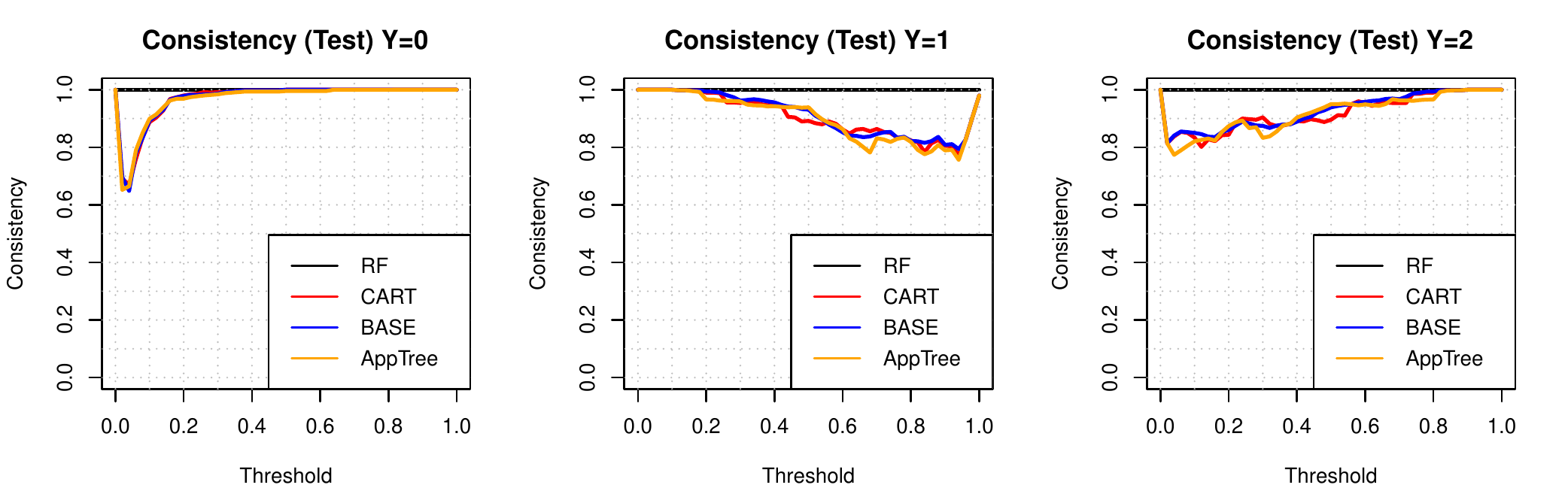}
   \includegraphics[width=6in]{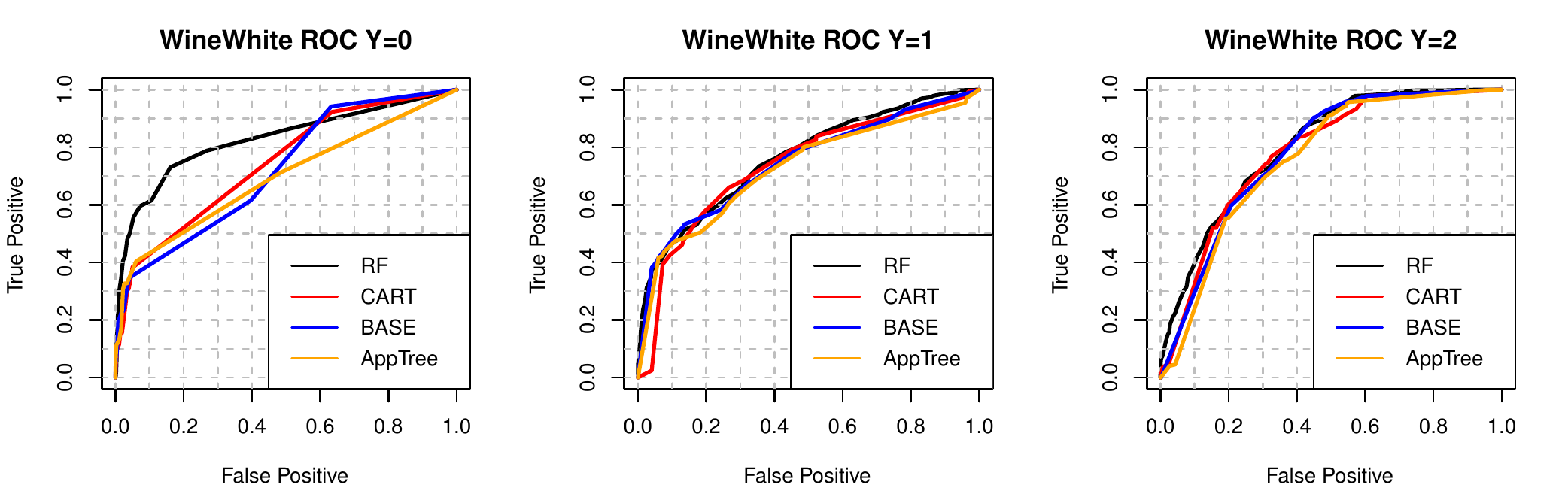}
   \includegraphics[width=6in]{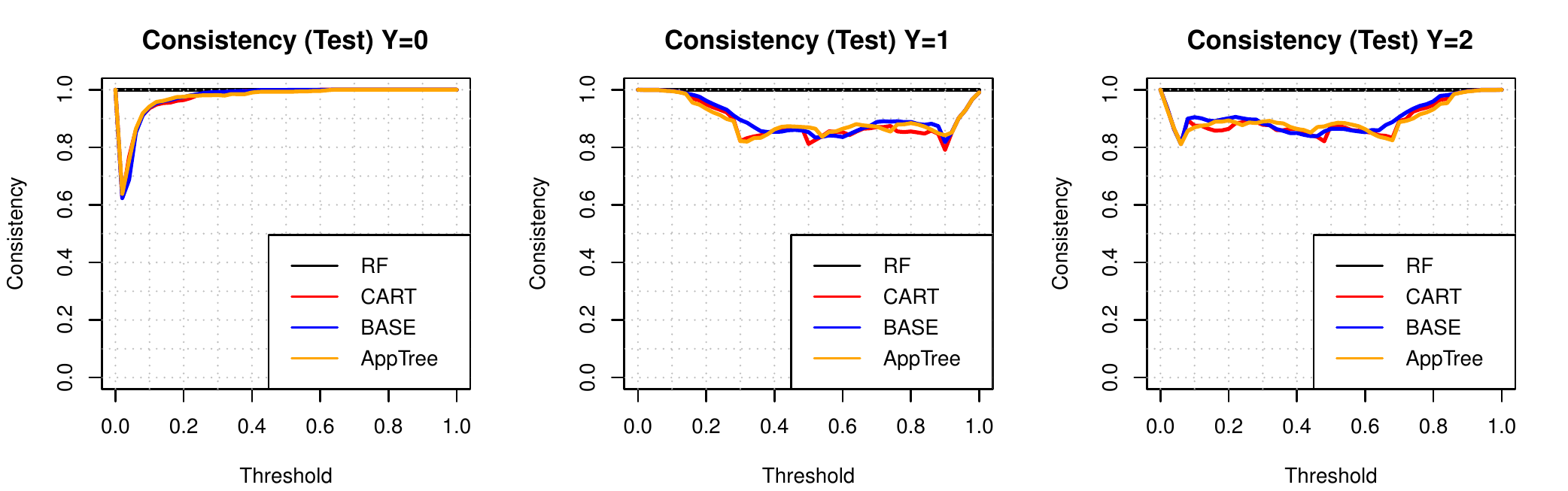}
   \caption{Performance evaluation on multiclass (3-class) classification datasets WineRed (top two rows) and WineWhite (bottom two rows). ROC curves are plotted in a one v.s. all fashion. Consistency is only checked on testing data. }
   \label{fig:mi12}
\end{figure}

Figures \ref{fig:mi11} and \ref{fig:mi12} demonstrate evaluate performance on the three multiclass classification datasets. We observe similar patterns as we did in binary cases that all three tree methods work similarly. ROC curves of RF show poorer performance than in binary classification, and ROC curves of the three tree methods are further from random forests, suggesting that deeper trees might be necessary. Nonetheless, all tree methods again agree with the RF on about 80\% of the predictions made by RF on the test data. We have therefore shown that our stability request of AppTree does not undermine its predictive power and consistency with the black box when compared with other tree methods.

\subsection{Stability}

\begin{figure}[htbp] 
   \centering
   \includegraphics[width=6in]{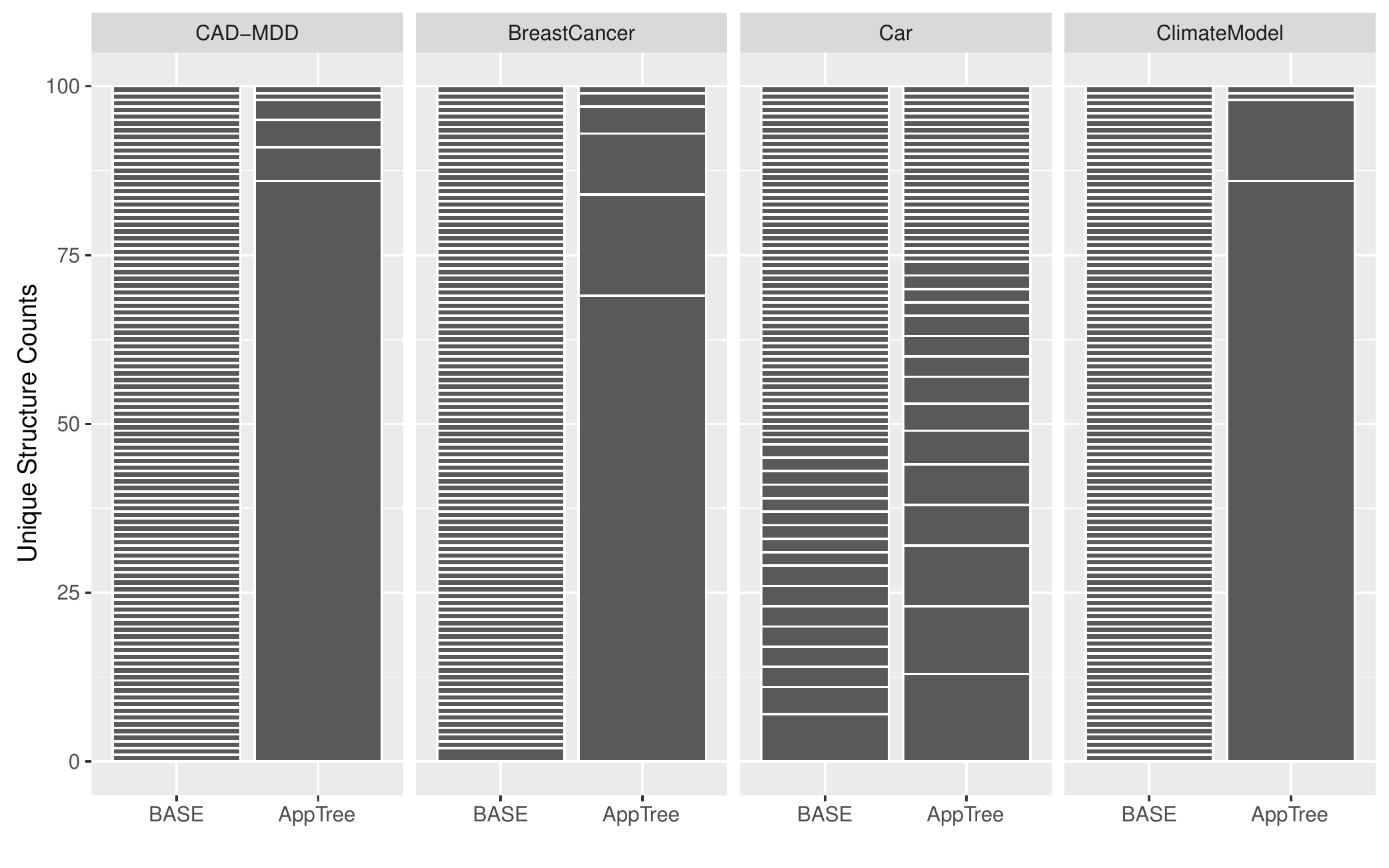}
   \caption{BASE and AppTree stability measured on binary classification datasets. From left to right: CAD-MDD, BreastCancer, Car, ClimateModel. In each column, a single black bar represents a unique structure of the tree, while the height of the bar represents the number of occurrence of that structure out of 100 replications.}
   \label{fig:strbi1}
\end{figure}

\begin{figure}[htbp] 
   \centering
   \includegraphics[width=6in]{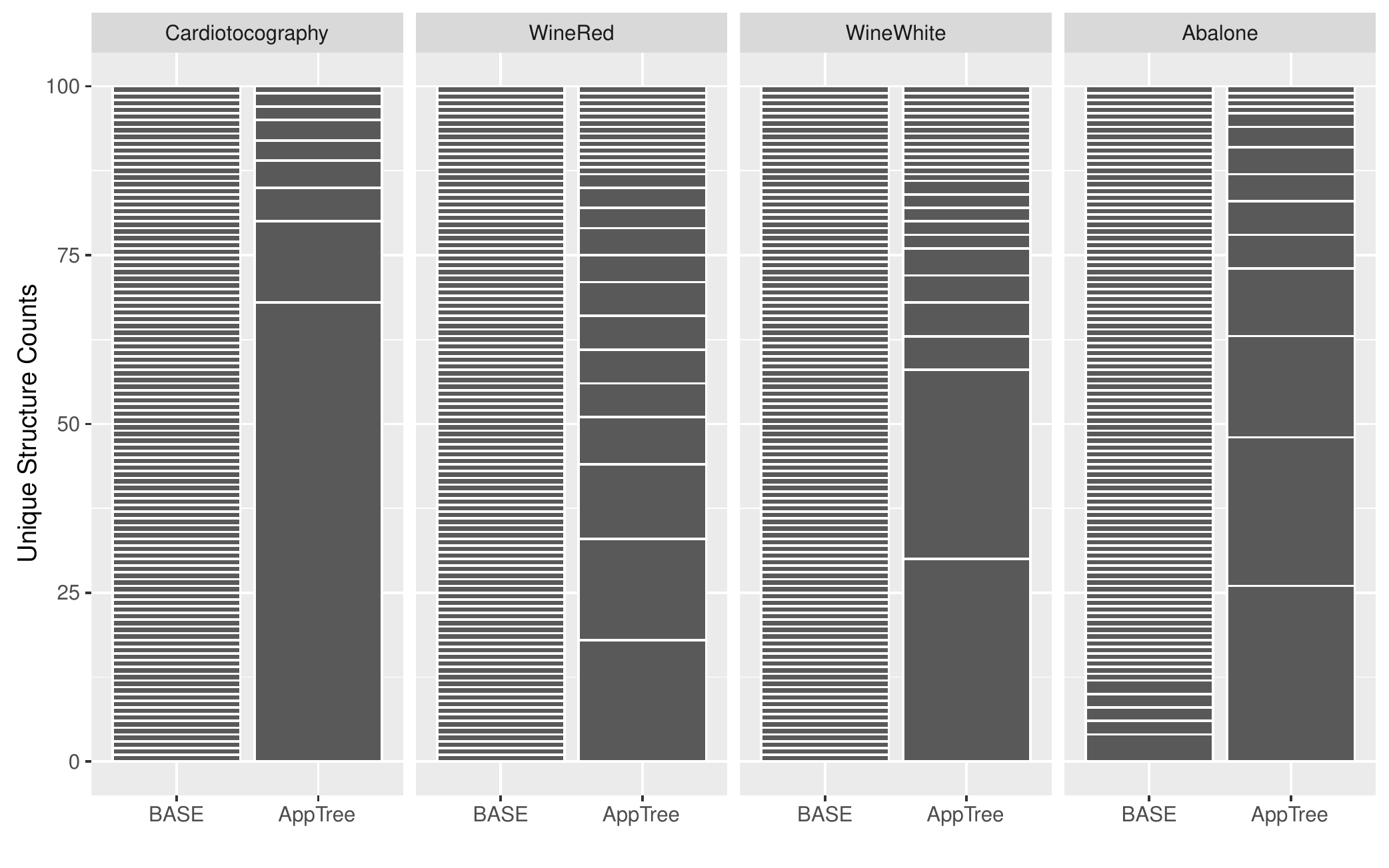}
   \caption{BASE and AppTree stability measured on multiclass classification datasets. From left to right: Cardiotocography, WineRed, WineWhite, Abalone. In each column, a single black bar represents a unique structure of the tree, while the height of the bar represents the number of occurrence of that structure out of 100 replications.}
   \label{fig:strmi1}
\end{figure}

Stability is the major concern in our simulation study. We measure how many different tree structures BASE and AppTree report out of their 100 replications of approximating the same RF, and count how many times each individual tree structure (both splitting covariate and splitting value) occurs. Table \ref{tab:stable} shows the number of different tree structures and number of occurrences of the top 3 frequent structures for both BASE and AppTree on each dataset. Figure \ref{fig:strbi1} and \ref{fig:strmi1} visualize these results.

\begin{table}[h]
   \centering
   \begin{tabular}{ccccc} 
      \toprule
      & \multicolumn{2}{c}{BASE} & \multicolumn{2}{c}{AppTree} \\
      Name & \#Struct & Top 3 Cnt & \#Struct & Top 3 Cnt\\
      \midrule
      CAD-MDD & 100 &1, 1, 1 & 6 & {\bf 86}, 5, 4\\
      BreastCancer & 99 & 2, 1, 1& 6 & {\bf 69}, 15, 9\\
      Car & 70 & 7, 4, 3& 41& {\bf 13}, 10, 9\\
      ClimateModel & 100 & 1, 1, 1 & 4 &{\bf 86}, 12, 1\\
      Abalone & 93 & 4, 2, 2 & 14 & {\bf 26}, 22, 15 \\
      Cardiotocography & 100 & 1, 1, 1 & 9 & {\bf 68}, 12, 5\\
      WineRed & 100 & 1, 1, 1 & 26 & {\bf 18}, 15, 11\\
      WineWhite & 100 & 1, 1, 1 & 25 & {\bf 30}, 28, 5\\
      \bottomrule
   \end{tabular}
    \caption{Stability of BASE and AppTree. The table shows the number of identical structures out of 100 replications and counts the occurrences of the top 3 structures in each case. Boldfaced numbers show the occurrences of the dominant tree structure out of 100 replications generated by AppTree for each dataset.}
   \label{tab:stable}
\end{table}

BASE is a non-adaptive version of AppTree that only requests pseudo samples once at the root node. Our simulation setting guarantees that BASE and AppTree have access to the same set of all possible splitting covariates and values. If we compared AppTree with BASE equipped with an enormous amount of pseudo samples at the beginning such that at each node BASE had no fewer sample points than AppTree, we should expect similar behavior between those two methods. However, BASE fails to stabilize the tree structure in our experiment as almost every 6-layer tree it produces has unique structure, whereas AppTree manages to generate a small number of dominant tree structures with a confidence control of $\alpha=0.1$. Thus, our adaptive increment of the pseudo sample size significantly contributes to the stability of the decision tree we obtain from the coaching procedure as the approximation tree.

Notice that $0.95^{31} \approx 0.2$, which means if we choose $\alpha=0.05$ and train with with infinitely many pseudo samples, we should have the most dominant 6-layer tree structure occurring about 20 out of 100 replications. Our results on most of the datasets have already attained such stability with $\alpha=0.1$ and $N_{ps} = 5\times 10^5$, therefore the control of $\alpha$ is relative conservative while the choice of the pseudo sample cap $N_{ps}=5 \times 10^5$ is sufficient. The significance level $\alpha$ controls the stability at a split-wise level. It is possible to extend this to further stabilize the tree by again introducing the FWER at the tree level. Notice this procedure may also increase the number of pseudo samples we need at each split.

\section{Approximation Tree for CAD-MDD Data}
\label{app:mdd}

See table \ref{mdd} for full details of the approximation tree for CAD-MDD data.
\begin{table}[]
\centering
\begin{tabular}{|c|c|c|c|}
\hline
Covariate & Value & Sample\_Number & p \\ \hline 
46        & 1.5   & 836            & 4.3e-158 \\ \hline 
53        & 1.5   & 508            & 1.8e-22  \\ \hline 
39        & 1.5   & 474            & 3.3e-08   \\ \hline 
54        & 0.5   & 427            & 9.9e-06  \\ \hline 
40        & 2.5   & 47             & 2.3e-18  \\ \hline 
48        & 2.5   & 34             & 3.0e-13  \\ \hline 
17        & 1.5   & 28             & 4.8e-14  \\ \hline 
14        & 1.5   & 318            & 3.8e-34  \\ \hline 
50        & 2.5   & 101            & 2.7e-13  \\ \hline 
58        & 1.5   & 70             & 4.7e-10  \\ \hline 
27        & 1.5   & 31             & 8.7e-10  \\ \hline 
48        & 2.5   & 227            & 1.7e-18  \\ \hline 
53        & 1.5   & 75             & 2.0e-14  \\ \hline 
35        & 2.5   & 152            & 2.3e-06 \\ \hline 
\end{tabular}
\caption{Details of the approximation tree for the CAD-MDD Data. }\label{mdd}
\end{table}

\section{Approximation Tree for COMPAS Data}
\label{app:imp}

The original dataset we downloaded was \texttt{https://github.com/propublica/compas-analysis/blob/master/compas-scores-two-years.csv}. 
We only keep eight features (sex, age, race, juv\_fel\_count, juv\_misd\_count, \_other\_count, priors\_count, c\_charge\_degree) and one target value (is\_recid). There are 7214 training data in total.

See table \ref{compas} for full details of the approximation tree for COMPAS data.

\begin{table}[]
\centering
\begin{tabular}{|c|c|c|c|}
\hline
Covariate & Value & Sample\_Number & p   \\ \hline 
priors\_count   & 2.5   & 7214           & 3.8e-22 \\ \hline 
age             & 23.5  & 4387           & 2.1e-46 \\ \hline 
priors\_count   & 0.5   & 981            & 7.0e-07 \\ \hline 
age             & 21.5  & 512            & 3.2e-02 \\ \hline 
is\_male        & 0.5   & 469            & 4.2e-16 \\ \hline 
age             & 31.5  & 3406           & 1.3e-08 \\ \hline 
priors\_count   & 1.5   & 1400           & 1.4e-04 \\ \hline 
priors\_count   & 1.5   & 2006           & 6.0e-02 \\ \hline 
age         & 33.5  & 2827           & 6.2e-08 \\ \hline 
priors\_count   & 4.5   & 1460           & 4.9e-05 \\ \hline 
age         & 27.5  & 542            & 7.4e-04 \\ \hline 
is\_African\_American & 0.5   & 918            & 1.5e-02 \\ \hline 
priors\_count       & 5.5   & 1267           & 3.2e-02 \\ \hline 
age         & 36.5  & 571            & 2.1e-02 \\ \hline 
priors\_count       & 8.5   & 796            & 3.2e-03 \\ \hline 
\end{tabular}
\caption{Details for approximation tree for the COMPAS Data.  }\label{compas}
\end{table}

\end{document}